\documentclass[runningheads]{llncs}
\usepackage{graphicx}
\usepackage{color}
\usepackage{xcolor}
\usepackage[export]{adjustbox}

\usepackage{caption}
\usepackage{subcaption}
\usepackage{comment}
\usepackage{algorithm}
\usepackage{algorithmic}
\usepackage{amsmath}
\usepackage{amsfonts}
\usepackage{amssymb}
\usepackage{multirow}
\usepackage{tabularx,booktabs}
\usepackage[font=small,labelfont=bf]{caption}
\usepackage[pagebackref=true,breaklinks=true,letterpaper=true,colorlinks,bookmarks=false]{hyperref}
\usepackage{makecell}
\setcellgapes{3pt}
\usepackage{microtype}
\usepackage[misc,geometry]{ifsym}
\DeclareMathOperator{\avg}{avg}

\usepackage{ulem}
\begin{document}
	\title{Image Collation: \\Matching illustrations in manuscripts}
	\titlerunning{Image Collation}
	%
	\author{Ryad Kaoua\inst{1} \and Xi Shen\inst{1} \and
		Alexandra Durr\inst{2} \and Stavros Lazaris\inst{3} \and David Picard\inst{1} \and Mathieu Aubry\inst{1} (\Letter)}
	\authorrunning{Kaoua et al.}
	%
	\institute{ LIGM, Ecole des Ponts, Univ. Gustave Eiffel, CNRS, Marne-la-Vallée, France \and
		Université de Versailles-Saint-Quentin-en-Yvelines, France \and
		CNRS (UMR 8167), France}
	\maketitle              
	\begin{abstract}
		Illustrations are an essential transmission instrument. For an historian, the first step in studying their evolution in a corpus of similar manuscripts is to identify which ones correspond to each other. This image collation task is daunting 
		for manuscripts separated by many lost copies, spreading over centuries, which might have been completely re-organized and greatly modified to adapt to novel knowledge or belief and include hundreds of illustrations. Our contributions in this paper are threefold. 
		First, we introduce the task of illustration collation and a large annotated public dataset to evaluate solutions, including 6 manuscripts of 2 different texts with more than 2 000 illustrations and 1 200 annotated correspondences. 
		Second, we analyze state of the art similarity measures for this task and show that they succeed in simple cases but struggle for large manuscripts when the illustrations have undergone very significant changes and are discriminated only by fine details. Finally, we show clear evidence that significant performance boosts can be expected by exploiting cycle-consistent correspondences. Our code and data are available on \url{http://imagine.enpc.fr/~shenx/ImageCollation}.
		
	\end{abstract}
	
	%

	%
	
	\section{Introduction}
	\label{intro}
	\begin{figure}[t]
		\begin{center}
			\begin{subfigure}{\linewidth}
				\includegraphics[width=2.25cm,height=2.8cm]{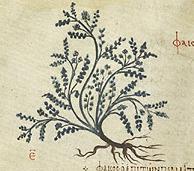}
				\includegraphics[width=2.25cm,height=2.8cm]{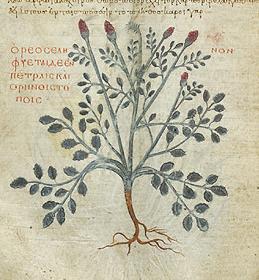}
				\includegraphics[width=2.25cm,height=2.8cm]{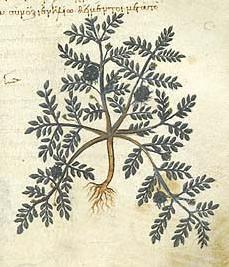}
				\includegraphics[width=2.25cm,height=2.8cm]{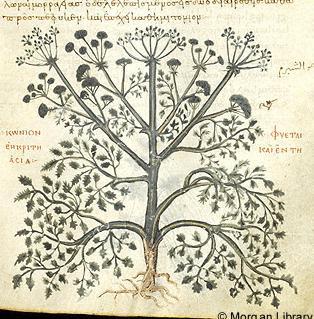}
				\includegraphics[width=2.25cm,height=2.8cm]{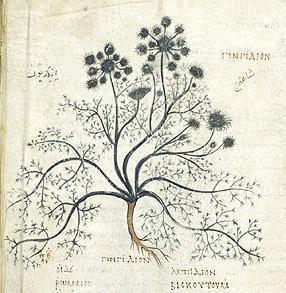}
				\caption{Illustrations of different plants from the same manuscript}\label{fig:teasera}
			\end{subfigure}
			\begin{subfigure}{\linewidth}
				\includegraphics[width=3.7cm,height=2.6cm]{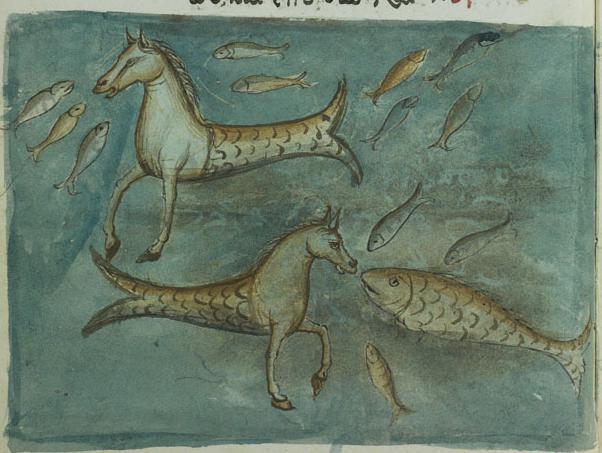}
				\includegraphics[width=3.7cm,height=2.6cm]{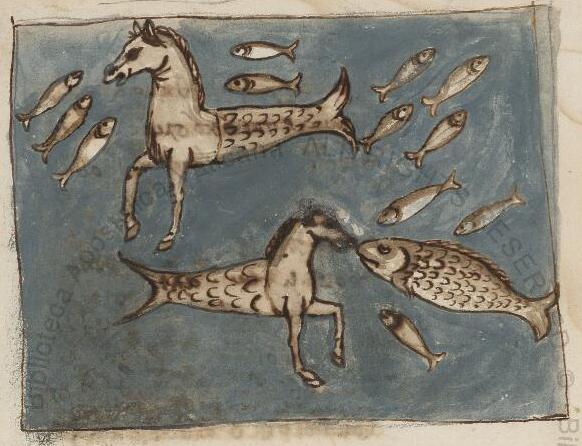}
				\includegraphics[width=3.7cm,height=2.6cm]{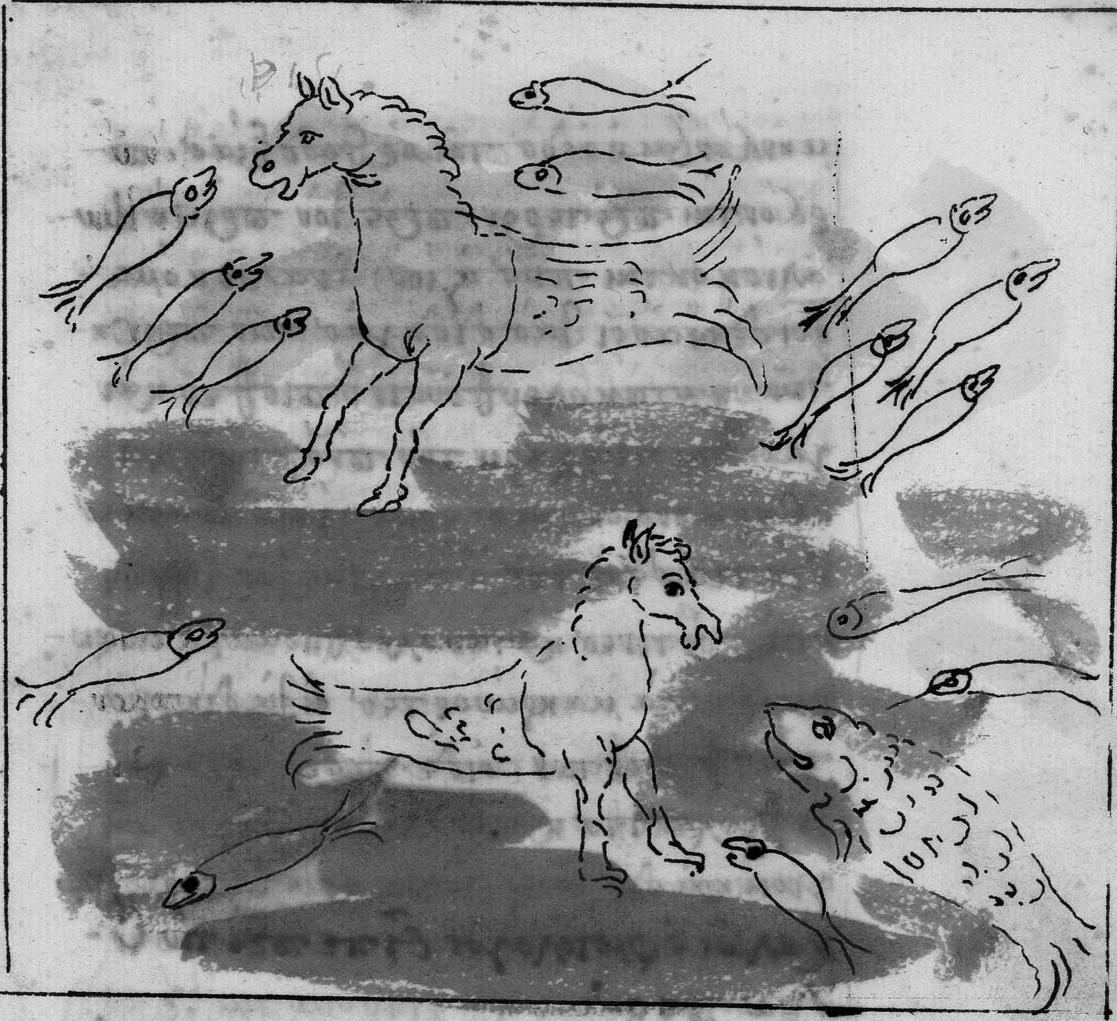}
				\caption{Same illustration in three  manuscripts with different styles}
				
				\label{fig:teaserb}
			\end{subfigure}
		\end{center}     
		\caption{The illustration alignment task we tackle is challenging for several reasons. It requires fine-grained separation between images with similar content~(a), while being invariant to strong appearance changes related to style and content modifications~(b). The examples presented in this figure are extracted from the two groups of manuscript we include in our dataset: (a) the {\it De Materia Medica} of Dioscoride; (b) the {\it Physiologus}.}
		\label{fig:teaser}
	\end{figure}
	
	Most research on the automatic analysis of manuscripts and particularly their alignment, also known as collation, has focused on text. However, illustrations are a crucial part of some documents, hinting the copyist values, knowledge and beliefs and are thus of major interest to historians. One might naively think that these illustrations are much easier to align than text and that a specialist can identify them in a matter of seconds. This is only true in the simplest of cases, where the order of the illustrations is preserved and their content relatively similar. In harder cases however, the task becomes daunting and is one of the important limiting factor for a large scale analysis.
	
	As an example, the \emph{``De materia Medica''} of Dioscorides, a Greek pharmacologist from the first century, has been widely distributed and copied between the 6th and the 16th century. Depending on the versions, it includes up to 800 depictions of natural substances. In particular the manuscripts we study in this paper contain around 400 illustrations of plants. They have been re-organized in different orders in the 17 different known illustrated versions of the text, for example alphabetically or depending on their therapeutic properties. The changes in the illustrations and their organizations hints both at the tradition from which each manuscript originates and at the evolution of scientific knowledge. However, the important shifts both in the illustrations appearance and in the order in which they appear makes identifying them extremely time consuming. While the text could help, it is not always readable and it is sometime not next to the illustrations.
	
	From a Computer Vision perspective, the task of retrieving corresponding illustrations in different versions of the manuscripts present several interesting challenges, illustrated in Figure~\ref{fig:teaser}. First, we are faced with a fine-grained problem, since many illustrations correspond to similar content, such as different plants (Figure~\ref{fig:teasera}). Second, the style, content and level of details vary greatly between different versions of the same text (Figure~\ref{fig:teaserb}). 
	Third, we cannot expect relevant supervision but can leverage many constraints. On the one hand the annotation cost is prohibitive and the style and content of the illustrations vary greatly depending on the manuscripts and topics. On the other hand the structure of the correspondences graph is not random and could be exploited by a learning or optimization algorithm. For example, correspondences should mainly be one on one, local order is often preserved, and if three or more versions of the same text are available correspondences between the different versions should be cycle consistent.

	In this paper, we first introduce a dataset for the task of identifying correspondences between illustrations of collections of manuscripts with more than 2 000 extracted illustrations and 1 200 annotated correspondences.

	Second, we propose approaches to extract such correspondences automatically, outlining the crucial importance both of the image similarity and its non-trivial use to obtain correspondences. Third, we present and analyze results on our dataset, validating the benefits of exploiting the problem specificity and outlining limitations and promising directions for future works.

	\begin{table}[t]
		\centering
		\caption{The manuscripts in our dataset come from two different texts, have diverse number of illustrations and come from diverse digitisations. In total, it includes more than 2000 illustrations and 1 200 annotated correspondences. } 
		\begin{tabular}{cccccc}
			\noalign{\smallskip}\hline\noalign{\smallskip}
			name & code~~ & number of~~ & folios'  & number of~~ & \multicolumn{1}{c}{annotated} \\
			& & folios & resolution~~ & illustrations~~ &\multicolumn{1}{c}{correspondences}\\
			\noalign{\smallskip}\hline\noalign{\smallskip}
			Physiologus & P1 & 109 & 1515x2045 & 51 & P2: 50 - P3: 50 \\
			" & P2 & 176 & 755x1068 & 51 & P1: 50 - P3: 51 \\
			" & P3 & 188 & 792x976 & 52 & P1: 50 - P2: 51 \\
			De Materia Medica & D1 & 557 & 392x555 & 816 & D2: 295 - D3: 524 \\
			" & D2 & 351 & 1024x1150 & 405 & D1: 295 - D3: 353 \\
			" & D3 & 511 & 763x1023 & 839 & D1: 524 - D2: 353 \\
			\noalign{\smallskip}\hline\\
		\end{tabular}
		\label{tab:manuscript_info}
		
	\end{table}

	\section{Related work}
	
	\paragraph{Text collation.}  The use of mechanical tools to compare different versions of a text can be dated back to Hinman’s collator, an opto-mechanical device which Hinman designed at the end of the 1940s to visually compare early impressions of Shaekspeare’s works~\cite{smith2000eternal}. More recently, computer tools such as CollateX~\cite{haentjens2015computer} have been developed to automatically compare digitised versions of a text. 
	The core idea is to explain variants using the minimum number of edits, or block move~\cite{bourdaillet2007practical}, to produce a variant graph~\cite{schmidt2009data}.
	Most text alignement methods rely on a transcription and tokenization step, which is not adapted to align images. 
	Methods which locally align texts and their transcriptions e.g.,~\cite{hobby1998matching,kornfield2004text,hassner2013ocr,sadeh2015viral,ezra2020transcription}, are also related to our task.
	Similar to text specific approaches, we will show that leveraging local consistency in the alignments has the potential to improve results for our image collation task.
	
	\paragraph{Image retrieval in historical documents.} Given a query image, image retrieval aims at finding images with similar content in a database.  Classic approaches such as Video-Google~\cite{sivic2003video} first look for images with similar SIFT~\cite{lowe2004distinctive} features then filter out those which features cannot be aligned by a simple spatial transformation. Similar bag of words approaches have been tested for pattern spotting in manuscripts~\cite{en2016scalable}. However, handcrafted features such as SIFTs fail in the case of strong style changes~\cite{shen2019discovering} which are characteristic of our problem.\\
	Recent studies~\cite{razavian2016visual,gordo2017end}
	suggest that directly employing global image features obtained with a network trained on a large dataset such as ImageNet~\cite{imagenet_cvpr09} is a strong baseline for image retrieval. Similarly, \cite{ubeda2019pattern} leverages features from a RetinaNet~\cite{lin2017focal} network trained on MS-Coco~\cite{lin2014microsoft} for pattern spotting in historical manuscripts and shows they improve over local features. If annotations are available, the representation can also to be learned specifically for the retrieval task using a metric learning approach, i.e., by learning to map similar samples close to each other and dissimilar ones far apart~\cite{gordo2017end,radenovic2018fine,revaud2019learning}. Annotations are however rare in the case for historical data. \\
	Recently, two papers have revisited the Video-Google approach for artistic and historical data using pre-trained deep local features densely matched in images to define an image similarity: \cite{shen2019discovering} attempts to discover repeated details in artworks collections and \cite{shen2019large} performs fine-grained historical watermark recognition. Both papers propose approaches to fine-tune the features in a self-supervised or weakly supervised fashion, but report good results with out-of-the box features.

	\section{Dataset and task}

	\begin{figure}[t]
		\centering
		\includegraphics[width=0.8\textwidth]{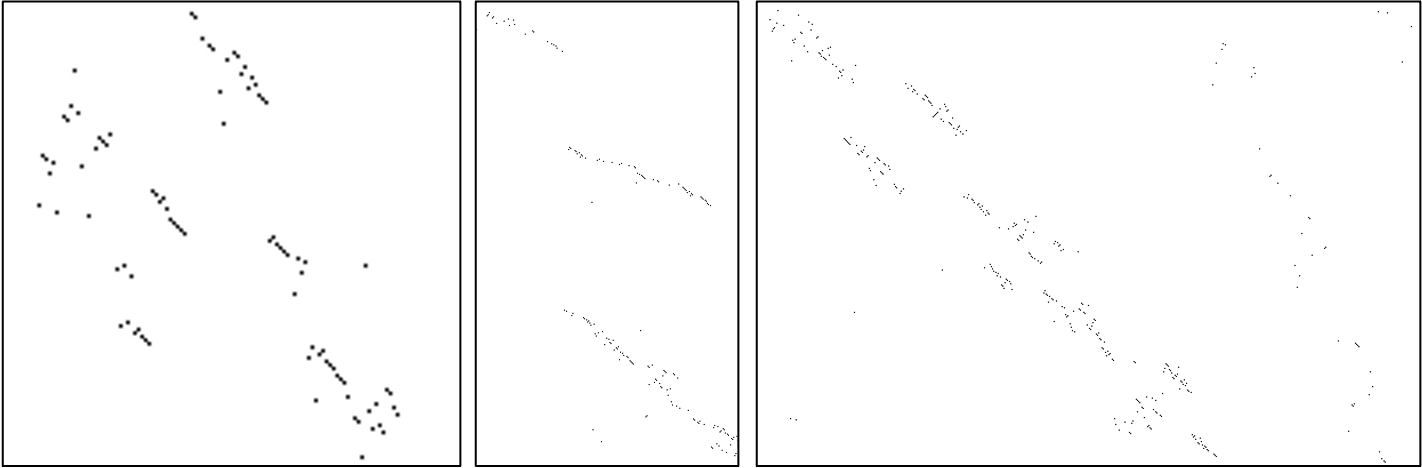}
		\caption{Structure of the correspondences for the "De Materia Medica". From left to right we show crops of the full correspondence matrices for D1-D2, D1-D3 and D2-D3. The black dots are the ground truth annotations. While the order is not completely random, the illustrations have been significantly re-ordered. Best viewed in electronic version.}
		\label{fig:order}
	\end{figure}
	
	\begin{figure*}[t]
		\centering     
		\begin{subfigure}{.49\linewidth}
			\centering
			\includegraphics[height=0.3\linewidth,width=0.3\linewidth]{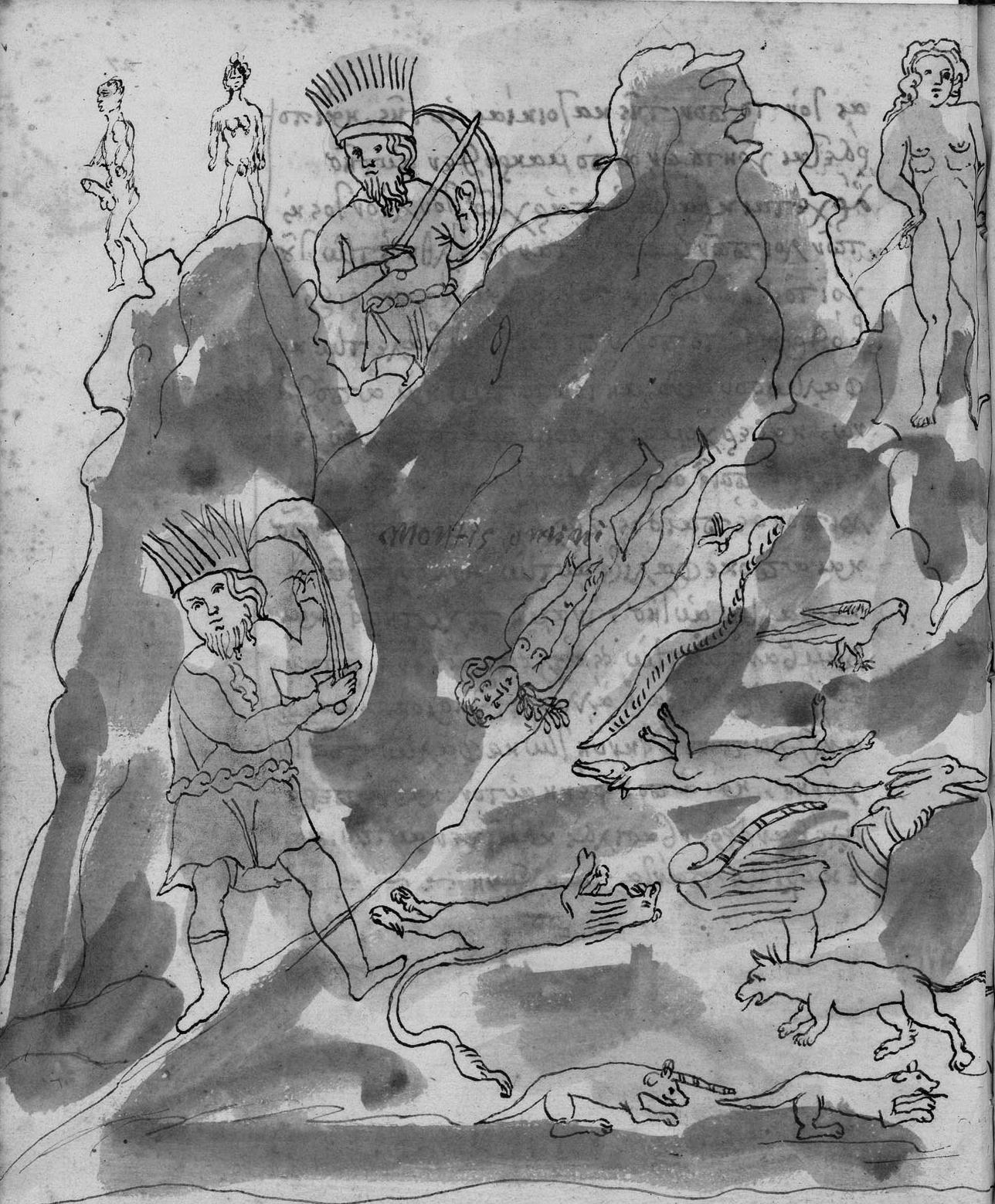}
			\includegraphics[height=0.3\linewidth,width=0.3\linewidth]{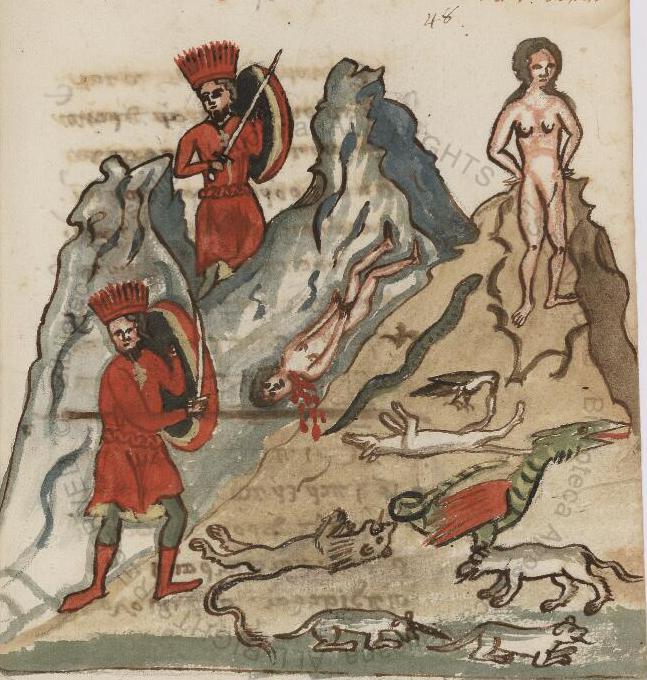}
			\includegraphics[height=0.3\linewidth,width=0.3\linewidth]{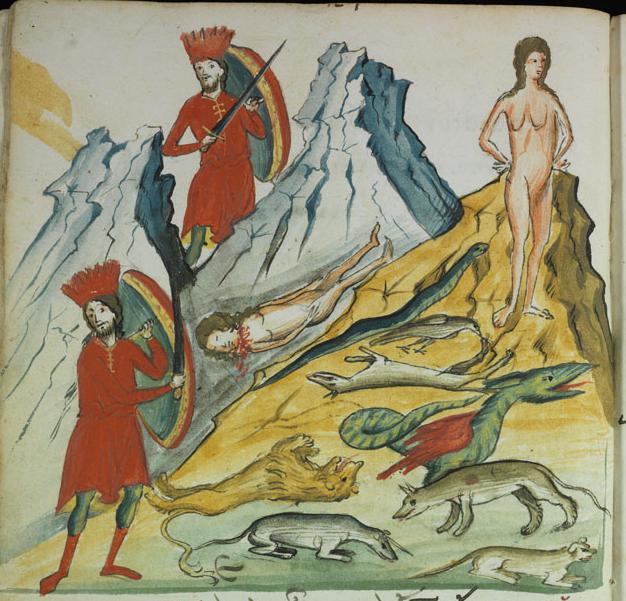}
			\includegraphics[height=0.3\linewidth,width=0.3\linewidth]{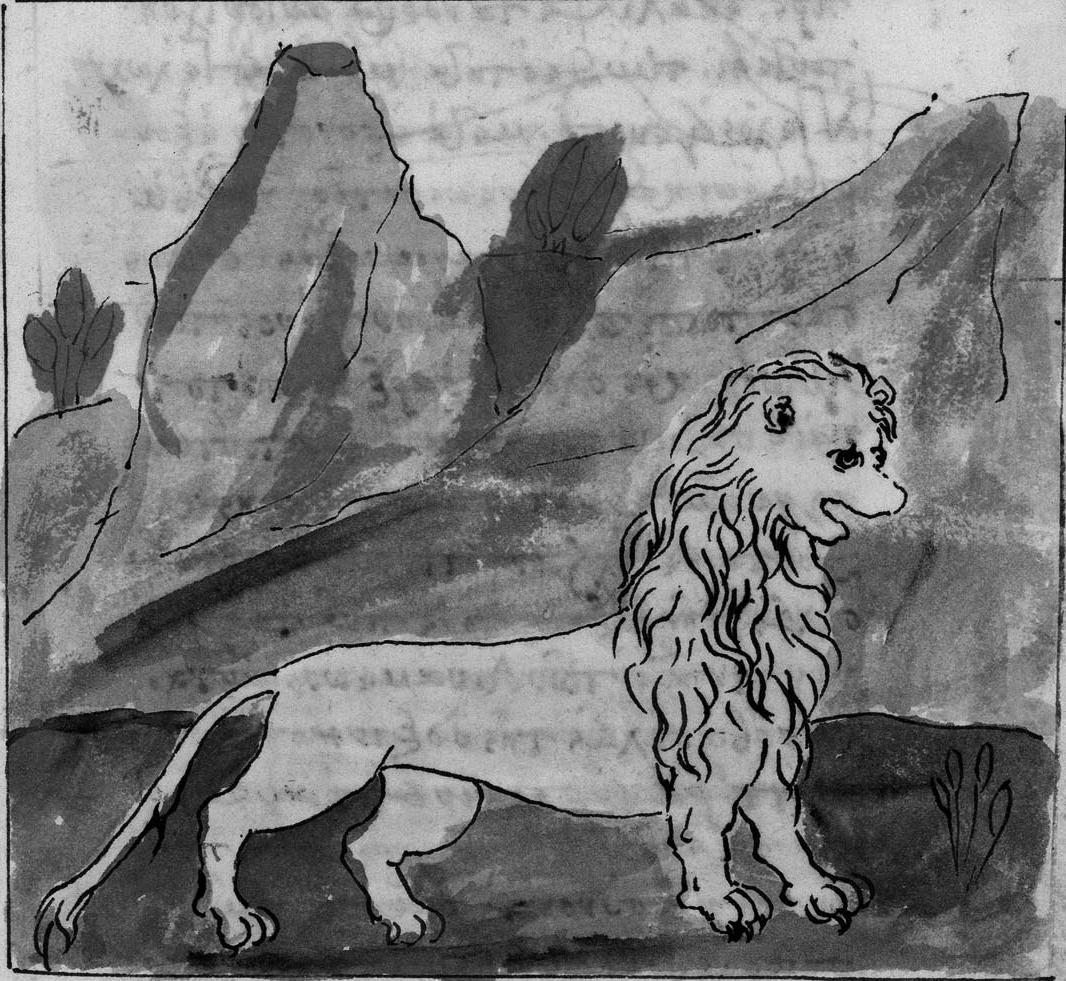}
			\includegraphics[height=0.3\linewidth,width=0.3\linewidth]{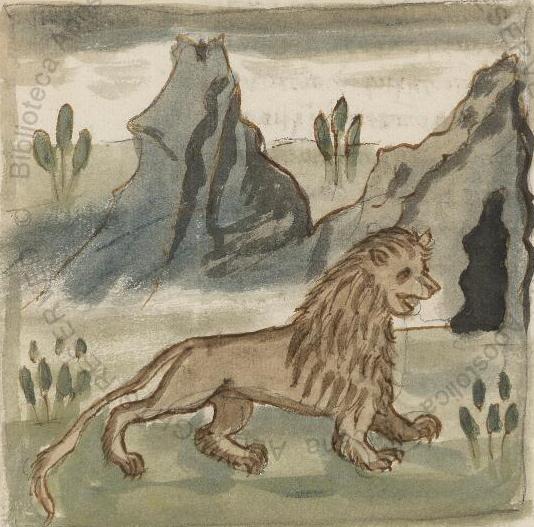}
			\includegraphics[height=0.3\linewidth,width=0.3\linewidth]{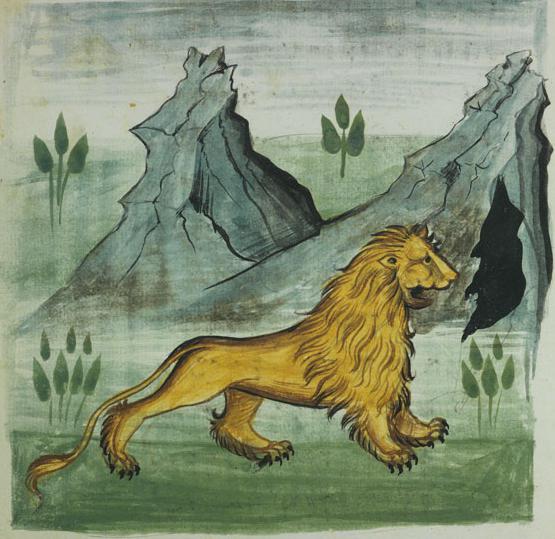}
			\includegraphics[height=0.3\linewidth,width=0.3\linewidth]{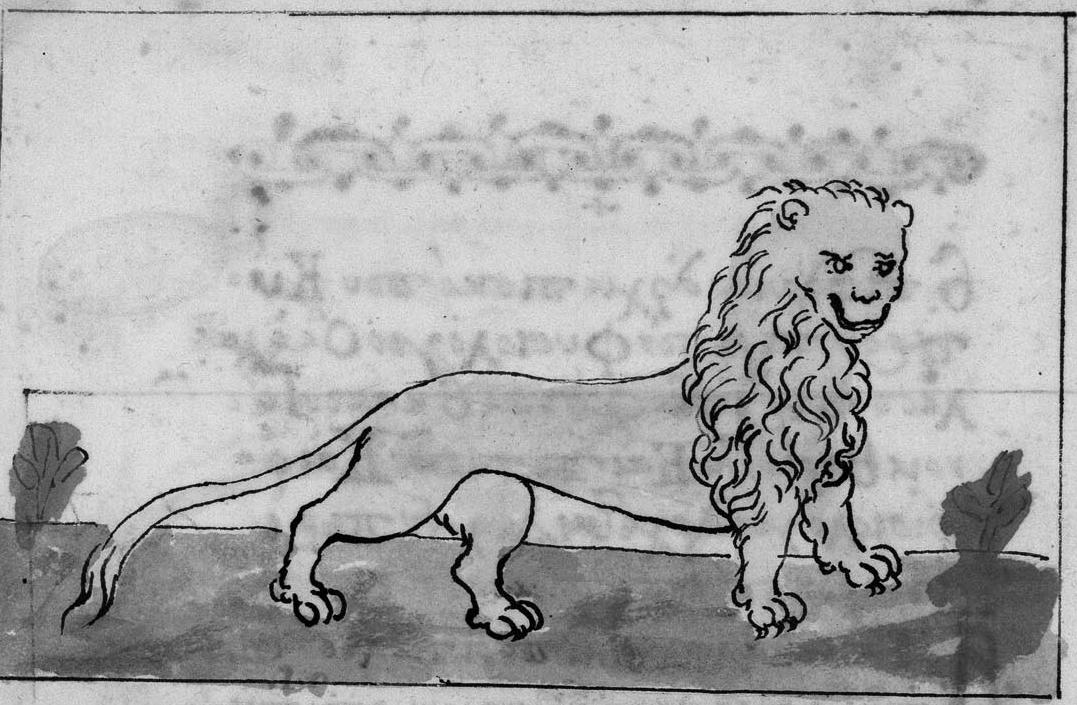}
			\includegraphics[height=0.3\linewidth,width=0.3\linewidth]{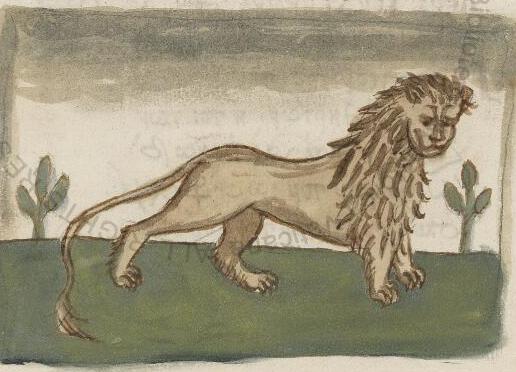}
			\includegraphics[height=0.3\linewidth,width=0.3\linewidth]{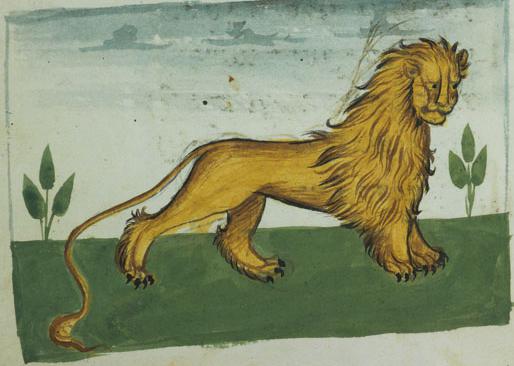}
			\caption{"Physiologus"}\label{fig:data_p}
		\end{subfigure}
		\begin{subfigure}{.49\linewidth}
			\centering
			\includegraphics[height=0.3\linewidth,width=0.3\linewidth]{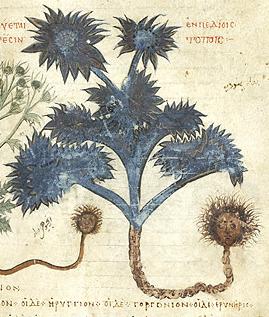}
			\includegraphics[height=0.3\linewidth,width=0.3\linewidth]{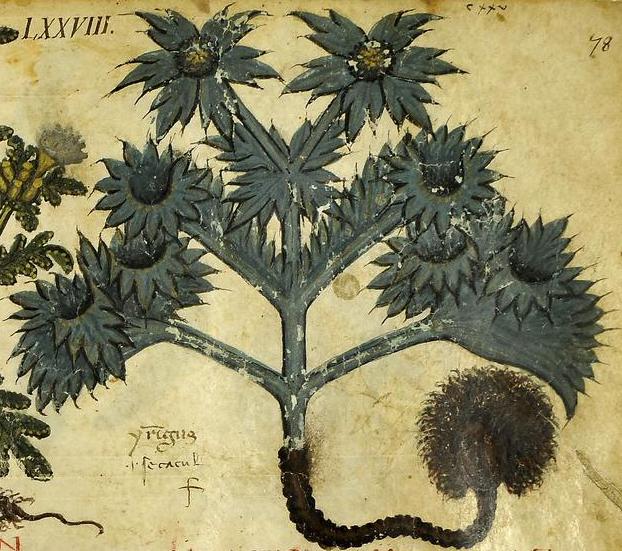}
			\includegraphics[height=0.3\linewidth,width=0.3\linewidth]{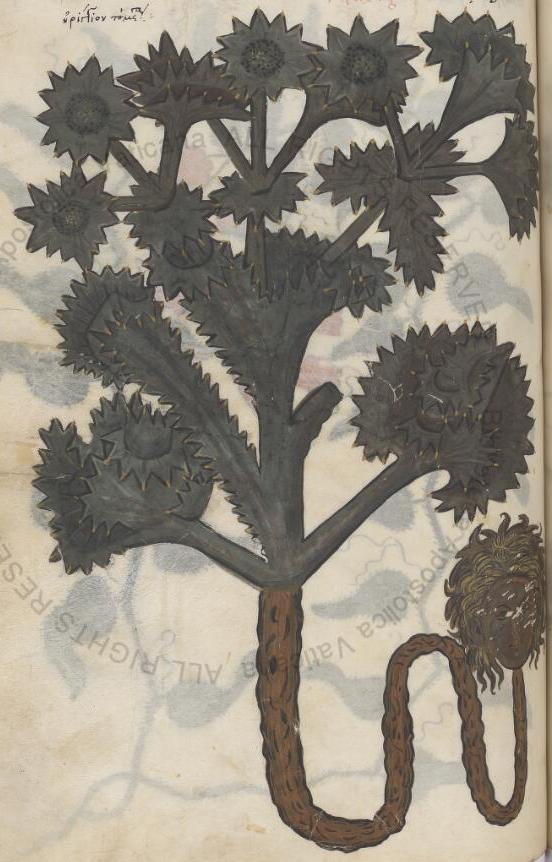}

			\includegraphics[height=0.3\linewidth,width=0.3\linewidth]{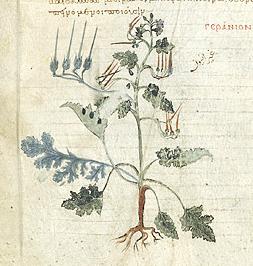}
			\includegraphics[height=0.3\linewidth,width=0.3\linewidth]{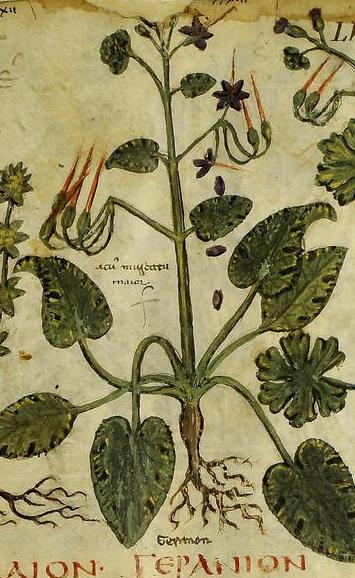}
			\includegraphics[height=0.3\linewidth,width=0.3\linewidth]{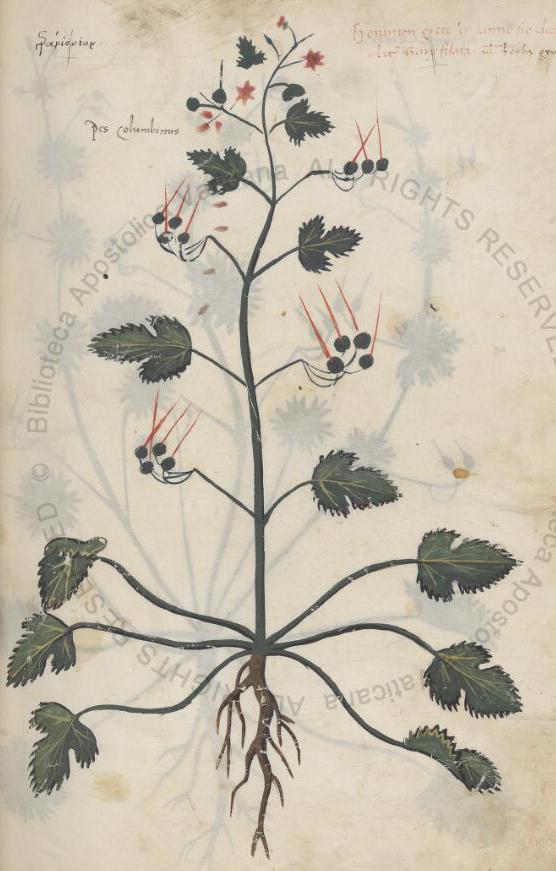}

			\includegraphics[height=0.3\linewidth,width=0.3\linewidth]{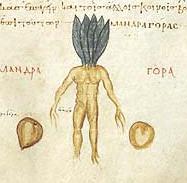}
			\includegraphics[height=0.3\linewidth,width=0.3\linewidth]{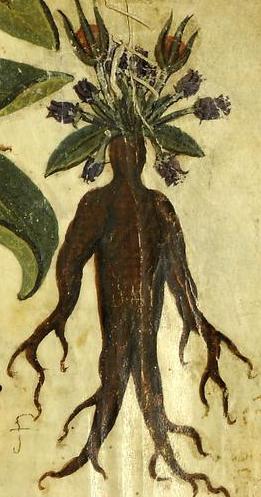}
			\includegraphics[height=0.3\linewidth,width=0.3\linewidth]{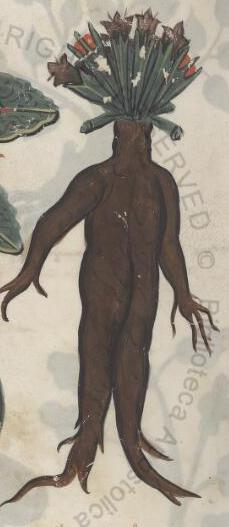}
			\caption{"De Materia Medica"}\label{fig:data_d}
		\end{subfigure}

		\caption{Examples of annotated image triplets in our two sets of manuscripts. Note how the depiction can vary significantly both in style and content.
		}
		\label{fig:data}
	\end{figure*}
	
	\label{sec:1}
	\label{sec:dataset}
	We designed a dataset to evaluate image collation, i.e., the recovery of corresponding images in sets of manuscripts.
	Our final goal is to provide a tool that helps historians analyze sets of documents by automatically extracting candidate correspondences. We considered two examples of such sets originating from different libraries with online text access to digitized manuscripts~\cite{wdl,themorgan,digitvatlib,internetculturale}, which characteristics are summarized in Table~\ref{tab:manuscript_info} and which are visualized in Figure~\ref{fig:teaser} and~\ref{fig:data}:
	\begin{itemize}
		\item The "Physiologus" is a christian didactic zoological text compiled in Greek during the 2nd century AD. The three manuscripts we have selected contain illuminations depicting real and fantastic animals and contain respectively 51, 51 and 52 illustrations which could almost all be matched between versions. In the three versions we used the order of the illustrations is preserved. The content of the illuminations is  similar enough that the correspondences could be easily identified by a human annotator. The style of the depictions however varied a lot as can be seen Figures~\ref{fig:teaserb} and~\ref{fig:data_p}.
		\item "De Materia Medica" was originally written in greek by Pedanius Dioscorides. The illustrations in the manuscript are mainly plants drawings. We consider three versions of the text with 816, 405 and 839 illustrations and annotated 295, 353 and 524 correspondences in the three associated pairs. Finding corresponding images in this set is extremely challenging due to three  difficulties: many plants are visually similar to each other (Figure~\ref{fig:teaserb}), the appearance can vary greatly in a matching image pair (Figure~\ref{fig:data_d}) and the illustrations are ordered differently in different manuscripts (Figure~\ref{fig:order}).
	\end{itemize}
	Note that we included three manuscripts in both sets so that algorithms and annotations could leverage cycle-consistency.
	
	\paragraph{Illustrations annotations.} We ran an automatic illustration extraction algorithm~\cite{monnier2020docExtractor} and found it obtained good results, but that some bounding boxes were inaccurate and that some different but overlapping illustrations were merged. To focus on the difficulty of finding correspondences rather than extracting the illustrations, we manually annotated the bounding boxes of the illustrations in each manuscript using the VGG Image Annotator~\cite{dutta2019vgg}. The study of joint detection and correspondence estimation is left for future work. 
	
	\paragraph{Correspondences annotations.} For the manuscripts of the Physiologus, annotating the corresponding illustration was time-consuming but did not present any significant difficulties. For the De Materia Medica however, the annotation presented significant challenge. Indeed, as explained above and illustrated in Figure~\ref{fig:order}, the illustrations have been significantly re-ordered, modified, and are often visually ambiguous. Since the manuscripts contain hundreds of illustrations, manually finding correspondences one by one was simply not feasible. We thus followed a three step procedure. First, for each illustration we used the image similarity described in Section~\ref{sec:method_score} to obtain its 5 nearest neighbors in each other manuscript. Second, we provided these neighbors and their context to a specialist who selected valid correspondences and searched neighboring illustrations and text to identify other nearby correspondences. Third, we used cycle consistency between the three manuscripts to validate the consistency of the correspondences identified by the specialist and propose new correspondences. 
	Interestingly, during the last step we noticed 51 cases where the captions and the depictions were not consistent. While worth studying from an historical perspective, these cases are ambiguous from a Computer Vision point of view, and we removed all correspondences leading to such inconsistencies from our annotations. 
	
	\paragraph{Evaluation metric. } We believe our annotations to be relatively exhaustive, however the difficulty of the annotation task made this hard to ensure. We thus focused our evaluation metric on precision rather than recall. More precisely, we expect algorithms to return a correspondence in each manuscript for each reference image and we compute the average accuracy on annotated correspondences only. In our tables, we report performances on pairs of manuscripts $M_1-M_2$ by finding correspondences in both directions (finding a correspondence in $M_2$ for each image of $M_1$, then a correspondence in $M_1$ for each image in $M_2$) and averaging performances.  Note that there is a bias in our annotations in the De Materia Medica since we initially provided the annotator with the top correspondences using our similarity. However, while it may slightly over-estimate the performance of our algorithm, qualitative analysis of the benefits brought by our additional processing remains valid.

	\section{Approach}
	\label{marker_approach_section}
\begin{figure*}[t]
		\centering

		\includegraphics[width=1\textwidth]{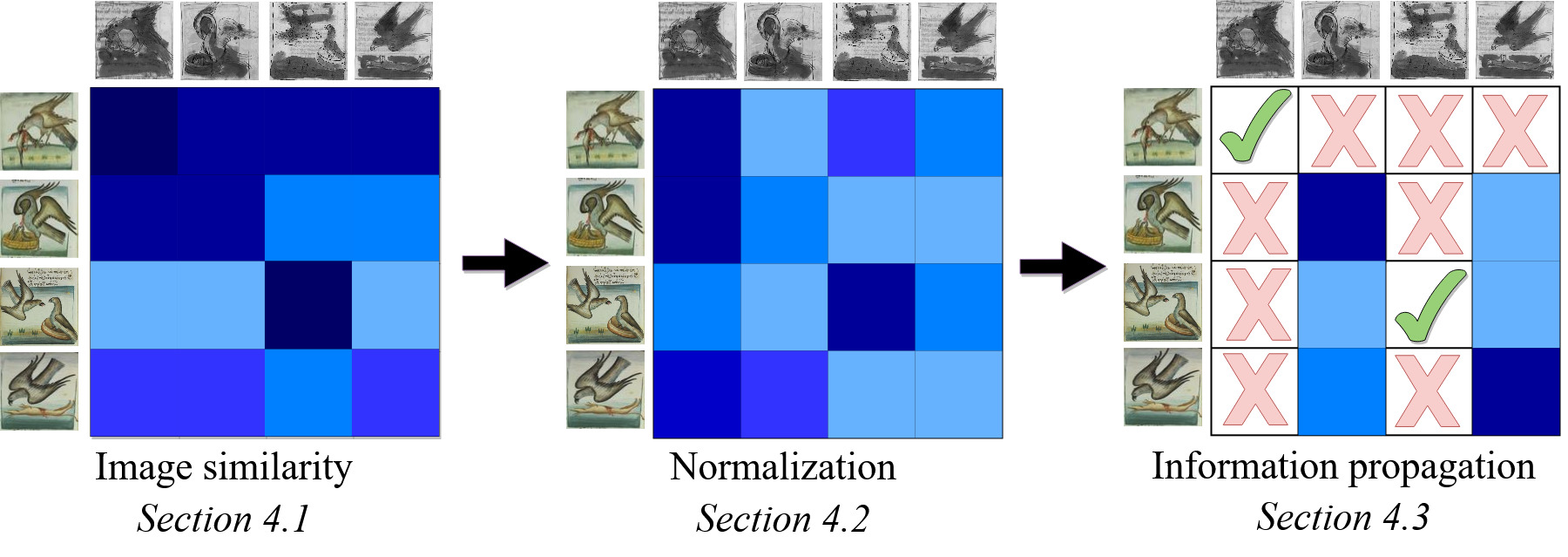}
		\caption{Overview of our approach. We first compute a similarity score between each pair of image, which we visualize using darker colors for higher similarity. We then normalize the similarity matrix to account for images that are similar to many other, such as the first line of our example. Finally, we propagate signal from confident correspondences which are maxima in both directions (green marks) to the rest of the matrix. }
		\label{fig:pipeline}
	\end{figure*}

	In this section, we present the key elements of our image collation pipeline, visualized in Figure~\ref{fig:pipeline}. {Except when explicitly mentioned otherwise,} we focus on studying correspondences in a pair of manuscripts. First, we discuss image similarities adapted to the task. Second, we introduce different normalizations of the similarity matrix associated to a pair of manuscripts. Third, we present a method to propagate information from confident correspondences to improve results. Finally, we give some implementation details.

	\subsection{Image similarities}
	\label{sec:sim}
	\label{sec:method_score}
	We focus on similarities based on deep features, following consistent observations in recent works that they improve over their classical counterparts for historical image recognition~\cite{shen2019discovering,ubeda2019pattern}. Since we want our approach to be directly applicable to new sets of images, with potentially very different characteristics, we use off-the-shelf features, without any specific fine-tuning. More precisely we used ResNet-50 ~\cite{he2016deep} features trained for image classification on ImageNet~\cite{imagenet_cvpr09}, which we found to lead to better performances (see Section~\ref{sec:res}). 
	
	\paragraph{Raw features.}
	Directly using raw features to compute image similarity is a strong baseline. Similar to other works~\cite{shen2019discovering,shen2019discovering}, we found that using {\it conv4} features and averaging cosine similarity of these features at the same location consistently performed best. More formally, given two images $I_1$ and $I_2$ we consider their {\it conv4} features $f_k=(f_k^i)_{i \in\left\lbrace 1,\ ...,\ N\right\rbrace}$, where $k=1$ or $2$ is the image ID, $i$ is the index of the spatial location in feature map and $N$ is the size of the feature map. We define the feature image similarity as:
	
	{\small\begin{equation}
		\begin{aligned}
		S_{features} (I_1, I_2) = \frac{1}{N} \sum_{i=1}^{N} \frac{f^i_1}{\|f^i_1\|} \cdot \frac{f^i_2}{\|f^i_2\|}
		\label{eqn:raw_feat}
		\end{aligned}
		\end{equation}}\normalsize
	where $\cdot$ is the scalar product. Note that the normalization is performed for each local feature independently and this similarity can only be defined if the two images are resized at a constant size. We used $256\times 256$ in our implementation.

	\paragraph{Matching-based similarity.}
	
	The feature similarity introduced in the previous paragraph only considers the similarity of local features at the same spatial location and scale, and not their similarity with other features at other locations in the image. 
	To leverage this information,~\cite{shen2019large} proposed to use a local matching score. Each feature $f^i_k$ of a source image $I_k$ is matched with the features extracted at several scales in a target images $I_l$. Then, each of the features of the target image is matched back in the source image and kept only if it matches back to the original feature, i.e. if it is a cycle consistent match. Finally, the best cycle consistent match among all scales of the target image  $m_{k,l}(f^i_k)$ is identified. Writing $x^i_k\in \mathbb{R}^2$ the position of the feature $f^i_k$ in the feature map and $x_{k,l}(f^i_k)\in \mathbb{R}^2$ the position of its best match $m_{k,l}(f^i_k)$ (which might be at a different scale), we define the similarity between $I_1$ and $I_2$ as:

	{\small\begin{equation}
		\begin{aligned}
		S_{matching}(I_1, I_2) =\frac{1}{2N}\sum_{i=1}^N 
		e^{-\frac{\|x^i_1-x_{1,2}(f^i_1)\|^2}{2\sigma^2} } \frac{f^i_1}{\|f^i_1\|}\cdot \frac{m_{1,2}(f^i_1)}{\|m_{1,2}(f^i_1)\|} \nonumber \\
		+\frac{1}{2N}\sum_{i=1}^N 
		e^{-\frac{\|x^i_2-x_{2,1}(f^i_2)\|^2}{2\sigma^2} } \frac{f^i_2}{\|f^i_2\|}\cdot \frac{m_{2,1}(f^i_2)}{\|m_{2,1}(f^i_2)\|}
		\label{eqn:local_match}
		\end{aligned}
		\end{equation}}\normalsize
	where $\cdot$ is the scalar product and $\sigma$ is a real hyperparameter.
	This score implicitly removes any contribution for non-discriminative regions and for details that are only visible in one of the depictions, since they will likely match to a different spatial location and thus have a very small contribution to the score. It will also be insensitive to local scale changes. 
	Note that~\cite{shen2019large} considered only the first term of the sum, resulting in a non-symmetric score. On the contrary, our problem is completely symmetric and we thus symmetrized the score.

	\paragraph{Transformation dependent similarity.}
	While the score above has some robustness to local scale changes, it assumes the images are coarsely aligned. To increase robustness to alignment errors, we follow~\cite{shen2019discovering} and use RANSAC~\cite{fischler1981random} to estimate a 2D affine transformation between the two images. More precisely, keeping the notations from the previous paragraph, we use RANSAC to find an optimal affine transformation $\mathcal{T}_{k,l}$ between image $I_k$ and $I_l$:
	{\small\begin{equation}\label{eqn:ransac_optimization}
		\mathcal{T}_{k,l}=\arg\max \sum_{i=1}^N 
		e^{-\frac{\|\mathcal{T}_{k,l}x^i_k-x_{k,l}(f^i_k)\|^2}{2\sigma^2} } \frac{f^i_k}{\|f^i_k\|}\cdot \frac{m_{k,l}(f^i_k)}{\|m_{k,l}(f^i_k)\|}
		\end{equation}}\normalsize
	Note this is slightly different from~\cite{shen2019discovering} which only uses the RANSAC to minimize the residual error in the matches to optimize the transformation. We found that maximizing the score instead of the number of inliers significantly improved the performances.
	Considering again the symmetry of the problem, this leads to the following score:
	{\small\begin{equation}
		\begin{aligned}
		S_{trans}(I_1, I_2) & =\frac{1}{2N}\sum_{i=1}^N 
		e^{-\frac{\|\mathcal{T}_{1,2}x^i_1-x_{1,2}(f^i_1)\|^2}{2\sigma^2} } \frac{f^i_1}{\|f^i_1\|}\cdot \frac{m_{1,2}(f^i_1)}{\|m_{1,2}(f^i_1)\|} \nonumber \\
		& +\frac{1}{2N}\sum_{i=1}^N 
		e^{-\frac{\|\mathcal{T}_{2,1}x^i_2-x_{2,1}(f^i_2)\|^2}{2\sigma^2} } \frac{f^i_2}{\|f^i_2\|}\cdot \frac{m_{2,1}(f^i_2)}{\|m_{2,1}(f^i_2)\|}
		\label{eqn:trans}
		\end{aligned}
		\end{equation}}\normalsize
	This score focuses on discriminative regions, is robust to local scale changes and affine transformations. We found it consistently performed best in our experiments, outperforming the direct use of deep features by a large margin.

	\subsection{Normalization} 
	\label{sec:norm}

	Let us call $S$ the similarity matrix between all pairs of images in the two manuscripts, $S(i,j)$ being a similarity such as the ones defined in the previous section 
	between the $i$th image of the first manuscript and the $j$th image of the second manuscript. For each image in the first manuscript, one can simply predict the most similar image in the second one as a correspondence, i.e. take the maximum over each row of the similarity matrix. This approach has however two strong limitations. First, it does not take into account that 
	some images tend to have higher similarity scores than other, resulting in rows or columns with higher values in the similarity matrix. Second, it does not consider the symmetry of the problem, i.e., that one could also match images in the second manuscript to images in the first one. 
	
	To account for these two effects, we propose to normalize the similarity matrix $S$ along each row and each column resulting in two matrices $R$ and $C$. 

	We experimented with five different normalization operations using softmax ($sm$), maximum ($\max$) and average (avg) operations either along the rows (leading to $R$) or the columns (leading to $C$), as shown in Table \ref{tab:eq_norm}.

	\begin{table}[t]
		\caption{Row-wise and Column-wise normalizations.}
		\tiny
		\centering{\makegapedcells
			\begin{tabular}{c|c|c}
				Normalization &   $R(i,j)$ &  $C(i,j)$ \\
				\hline
				$sm(\lambda S)$ & $\exp(\lambda S(i,j)) / \sum_k \exp(\lambda S(i,k))$ & $ \exp(\lambda S(i,j)) / \sum_k \exp(\lambda S(k,j))$ \\
				\hline
				$S/\avg(S)$ &  $R_{\avg}=S(i,j)/\avg_k S(i,k)$ & $C_{\avg}=S(i,j)/\avg_k S(k,i)$\\
				\hline
				$S/\max(S)$ &  ~$ R_{\max}=S(i,j)/\max_k S(i,k)$~ & ~$ C_{\max}=S(i,j)/\max_k S(k,i)$~ \\
				\hline 
				{ $sm(\lambda S/\avg(S))$ }& $ \exp(\lambda R_{\avg}(i,j)) / \sum_k \exp(\lambda R_{\avg}(i,k))$ & $ \exp(\lambda C_{\avg}(i,j)) / \sum_k \exp(\lambda C_{\avg}(k,j))$ \\
				\hline
				$sm(\lambda S/\max(s))$ & $ \exp(\lambda R_{\max}(i,j)) / \sum_k \exp(\lambda R_{\max}(i,k))$ & $ \exp(\lambda C_{\max}(i,j)) / \sum_k \exp(\lambda C_{\max}(k,j))$ \\
		\end{tabular}}
		\label{tab:eq_norm}
	\end{table}
	
	We then combine the two matrices $R$ and $C$ into a final score: we experimented with summing them or using element-wise (Hadamard) multiplication. Both performed similarly, with a small advantage from the sum, we thus only report those results.
	We found in our experiments that the $\max$ normalization performed best, without requiring an hyper-parameter. As such, our final normalized similarity matrix $N_S$ is defined as:
	{\small\begin{equation}
	\begin{aligned}
	N_S(i,j) = \frac{S(i,j)}{\max_k S(i,k)} + \frac{S(i,j)}{\max_k S(k,j)}
	\end{aligned}
	\end{equation}}\normalsize
	
	\subsection{Information propagation}
	\label{sec:prop}
	\label{marker_info_prop}

	While the normalized score $N_S$ obtained in the previous section includes information about both directions of matching in a pair of manuscripts, it does not ensure that correspondences are 2-cycle consistent, i.e. that the maxima in the rows of $N_S$ correspond to maxima in the columns. If one has access to more than 2 manuscripts, one can also check consistency between triplets of manuscripts and identify correspondences that are 2 and 3-cycle consistent. Correspondences that verify such cycle-consistency are intuitively more reliable, as we validated in our experiments, and thus can be used as anchors to look for other correspondences in nearby images. Indeed, while the order of the images is not strictly preserved in the different versions, there is still a clear locally consistent structure as can be seen in the ground truth correspondence matrices visualized in Figure~\ref{fig:order}.
	
	Many approaches could be considered to propagate information from confident correspondences and an exhaustive study is beyond the scope of our work. We considered a simple baseline as a proof of concept. Starting from an initial score $N_S$ and a set of confident correspondences $\mathcal{C}^\ast$ as seeds (e.g., correspondences that verify 2 or 3 cycle consistency constraints), we define a new score after information propagation $P_S$ as:
	
	{\small\begin{equation}
		\begin{aligned}
		P_S(i, j) =  N_S(i,j) \prod_{(k,l) \in \mathcal{C}^\ast} \left(1+\alpha \exp{\left(\frac{-||(i, j)-(k,l)||^2}{2\sigma_p^2}\right)}\right) 
		\end{aligned}
	\end{equation}}\normalsize
	where $\sigma_p$ and $\alpha$ are hyperparameters. Note that this formula can be applied with any definition of $\mathcal{C}^\ast$, and thus could leverage sparse correspondence  annotations.

	\begin{figure*}[t]
		\begin{center} 
			\begin{subfigure}{\linewidth}
				\includegraphics[width=0.15\linewidth,height=0.18\linewidth,cfbox=blue 1pt 1pt]{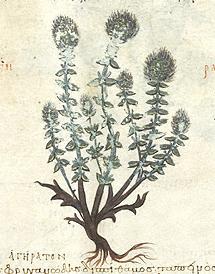}
				\includegraphics[width=0.15\linewidth,height=0.18\linewidth]{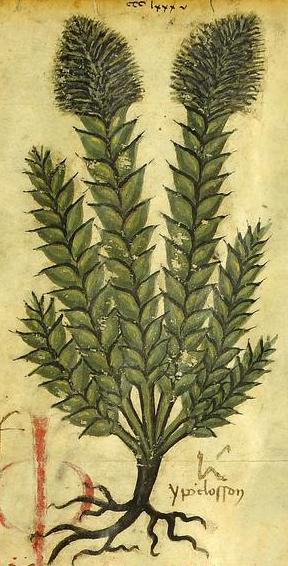}
				\includegraphics[width=0.15\linewidth,height=0.18\linewidth]{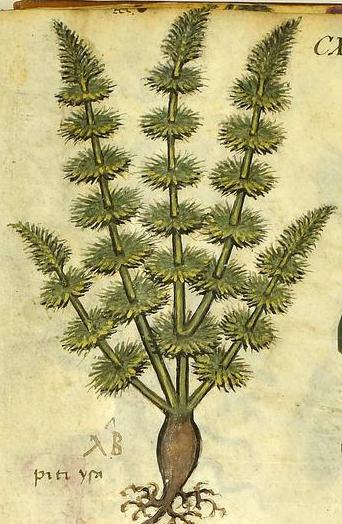}
				\includegraphics[width=0.15\linewidth,height=0.18\linewidth]{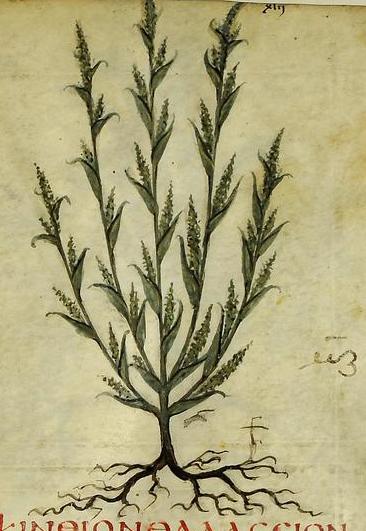}
				\includegraphics[width=0.15\linewidth,height=0.18\linewidth]{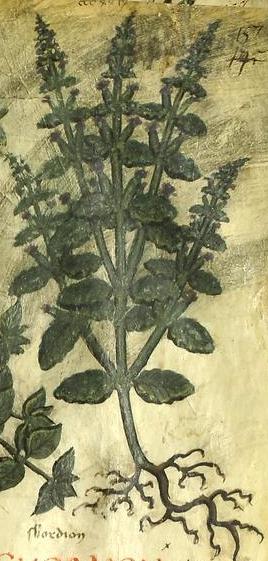}
				\includegraphics[width=0.15\linewidth,height=0.18\linewidth]{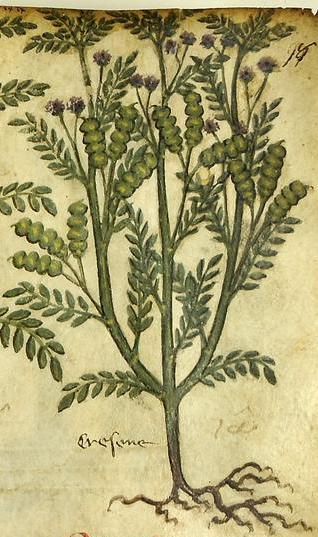}
				\caption{Query and 5 nearest neighbors according to our similarity score (Section \ref{sec:sim})}
			\end{subfigure}
			\begin{subfigure}{\linewidth}
				\includegraphics[width=0.15\linewidth,height=0.18\linewidth,cfbox=blue 1pt 1pt]{images/qual/r0.jpg}
				\includegraphics[width=0.15\linewidth,height=0.18\linewidth]{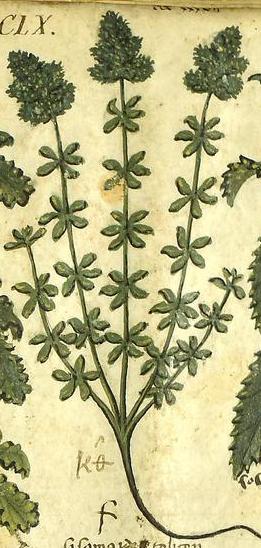}
				\includegraphics[width=0.15\linewidth,height=0.18\linewidth]{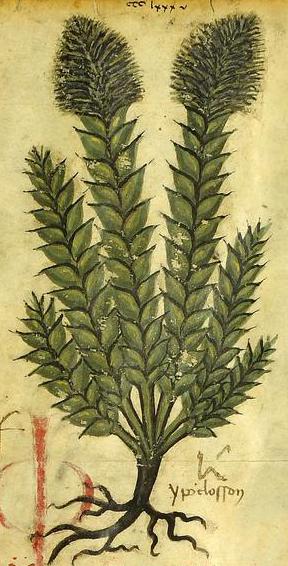}
				\includegraphics[width=0.15\linewidth,height=0.18\linewidth,cfbox=green 1pt 1pt]{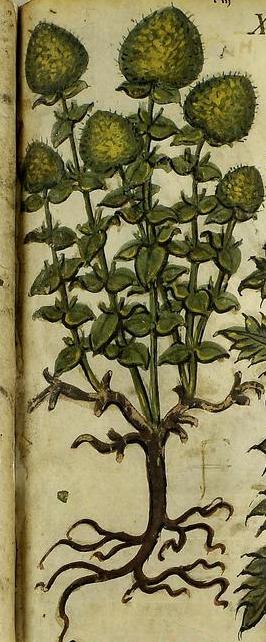}
				\includegraphics[width=0.15\linewidth,height=0.18\linewidth]{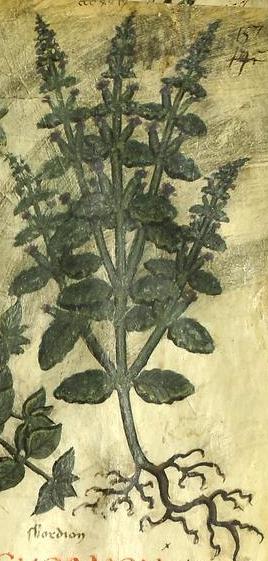}
				\includegraphics[width=0.15\linewidth,height=0.18\linewidth]{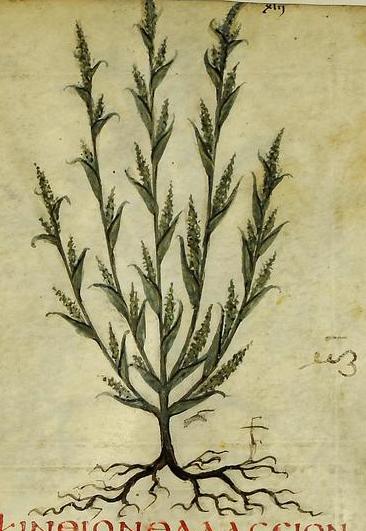}
				\caption{Query and 5 nearest neighbors according to our normalized score (Section \ref{sec:norm})}
			\end{subfigure}
			\begin{subfigure}{\linewidth}
				\includegraphics[width=0.15\linewidth,height=0.18\linewidth,cfbox=blue 1pt 1pt]{images/qual/r0.jpg}
				\includegraphics[width=0.15\linewidth,height=0.18\linewidth,cfbox=green 1pt 1pt]{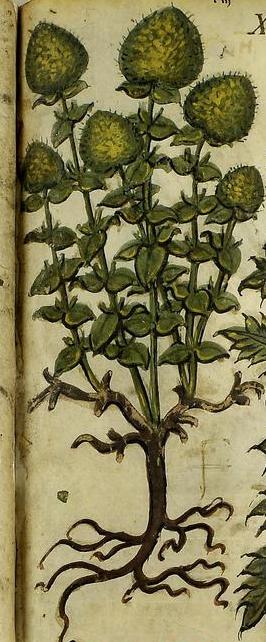}
				\includegraphics[width=0.15\linewidth,height=0.18\linewidth]{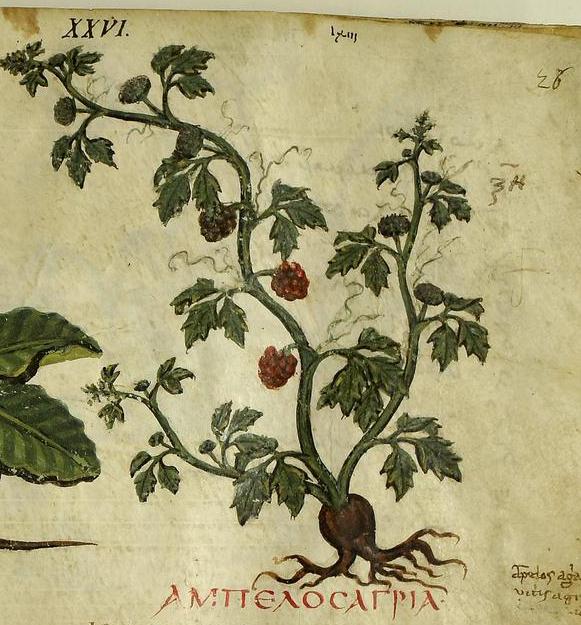}
				\includegraphics[width=0.15\linewidth,height=0.18\linewidth]{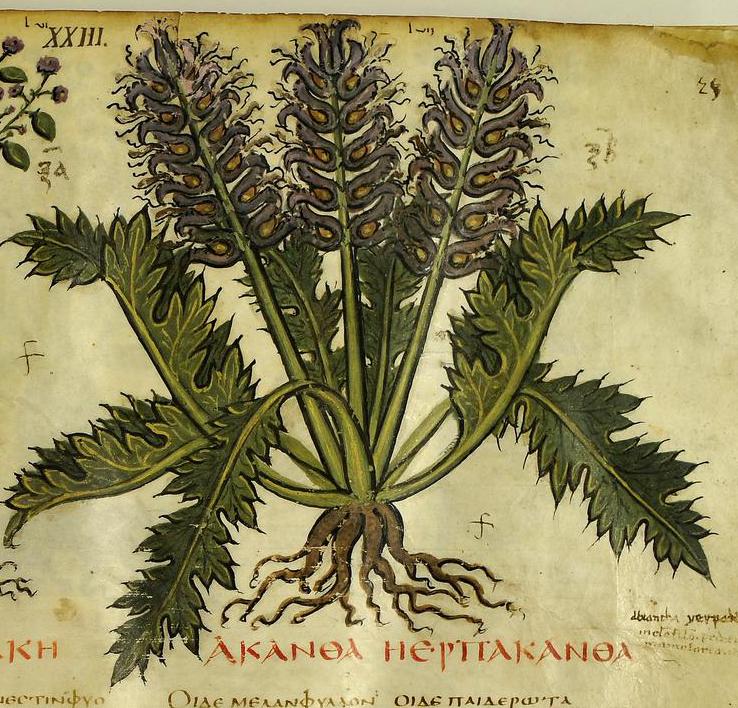}
				\includegraphics[width=0.15\linewidth,height=0.18\linewidth]{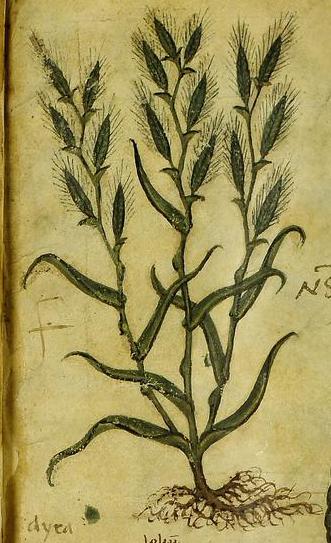}
				\includegraphics[width=0.15\linewidth,height=0.18\linewidth]{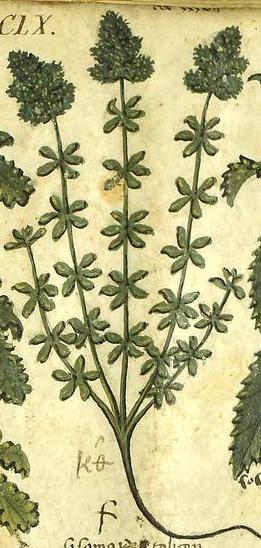}
				\caption{Query and 5 nearest neighbors after information propagation (Section \ref{sec:prop})}
			\end{subfigure}
		\end{center}   
		\caption{Query (in blue) and 5 nearest neighbors (ground truth in green) after the different steps of our method. Despite not being in the top-5 using the similarity, the correct correspondence is finally identified after the information propagation step}
		\label{fig:res}
	\end{figure*}

	\paragraph{Implementation details}
	
	In all the experiments, we extract {\it conv4} features of a ResNet50 architecture~\cite{he2016deep} pre-trained on ImageNet~\cite{imagenet_cvpr09}. To match illustrations between different scales, we keep the original aspect ratios and resize the source image to have 20 features in the largest dimension and the target image to five scales such that the numbers of features of the largest dimension are 18, 19, 20, 21, 22.
	We set $\sigma$ in Equation~\ref{eqn:local_match} and~\ref{eqn:trans} to $\frac{1}{\sqrt{50}}$ times the size of the image and the number of iterations in the RANSAC to 100. 
	For the information propagation, we find that $\sigma_p$ = 5 and $\alpha$ = 0.25 performs best. With our naive Pytorch implementation, computing correspondences between D1 (816 illustrations) and D2 (405 illustrations) takes approximately 80 minutes, 98\% being spent to compute similarities between all the 330,480 pairs of images.

	\section{Results}
	\label{sec:res}
	
	In this section, we present our results. 
	In~\ref{sec:res_sim} we compare different features similarities. In~\ref{sec:res_norm} we show the performance boost by the different normalizations. In~\ref{sec:res_ip} we demonstrate that the results can be improved by leveraging the structure of the correspondences. Finally, in~\ref{sec:res_lim} we discuss the failure cases and limitations.
	
	To measure the performance, as described in Section~\ref{sec:dataset}, we compute both the accuracy $a_1$ obtained by associating to each illustration of the first manuscript the illustration of the second manuscript which maximizes the score and the accuracy $a_2$ by associating illustrations of the second manuscript to illustrations of the first manuscript. We then report the average of these two accuracies $\frac{a_1+a_2}{2}$.
	Because using a good image similarity already led to almost perfect results on the Physiologus, we focus our analysis on the more challenging case of the De Materia Medica. The benefits of the three steps of our approach are illustrated in Figure~\ref{fig:res}, where one can also assess the difficulty of the task.
	
	\subsection{Feature similarity}	
	
	\begin{table}[t]
		\caption{ Percentage of accuracy of the correspondences obtained using $S_{features}$ with different conv4 features for all manuscripts pairs. 
		}
		\tiny
		\label{tab:conv4_score}
		\centering
		\begin{tabular}{c|c|c|c|c}
			Pairs ~& ResNet18~ & MoCo-v2~&  ArtMiner~&  ResNet50\\ \hline
			
			P1-P2~ & 78.0 & 75.0 & \bf 92.0 & 84.0 \\
			P1-P3~ & 75.0 & 62.0 & \bf 78.0 & 73.0 \\
			P2-P3~ & 99.0 & 99.0 & 98.0 & \bf 100.0 \\
			
			\noalign{\smallskip} 
		\end{tabular}
		~~~~
		\begin{tabular}{c|c|c|c|c}
			
			Pairs ~& ResNet18~ & MoCo-v2~&  ArtMiner~&  ResNet50\\ \hline
			
			D1-D2~ & 31.9 & 31.2 & 35.3 & \bf 35.4 \\
			D1-D3~ & 42.0 & 35.6 & 43.7 & \bf 46.1 \\
			D2-D3~ & 27.6 & 26.6 & 31.7 & \bf 34.1 \\
		
			\noalign{\smallskip} 
		\end{tabular}
	\end{table}
	
	\label{sec:res_sim}
	\begin{table}[t]
		\caption{Accuracy of the correspondences obtained using the different similarities explained in Section~\ref{sec:method_score}, as well as the similarity used in~\cite{shen2019discovering}, which is similar to $S_{trans}$ but uses the number of inliers instead of our score to select the best transformation.
		}
		\tiny
		\label{tab:score}
		\centering
		\begin{tabular}{c|c|c|c|c}
			Pairs ~& $S_{features}$~ & $S_{matching}$~&  ~\cite{shen2019discovering} ~&   $S_{trans}$\\ \hline
			P1-P2~ & 84.0 & 98.0 & 99.0 & \bf 100.0 \\
			P1-P3~ & 73.0 & 94.0 & \bf 98.0 & \bf 98.0 \\
			P2-P3~ & \bf 100.0 & 98.0 & \bf 100.0 & \bf 100.0 \\
			\noalign{\smallskip}
		\end{tabular}
		~~~~
		\begin{tabular}{c|c|c|c|c}
			Pairs ~& $S_{features}$~ & $S_{matching}$~&  ~\cite{shen2019discovering} ~&  $S_{trans}$\\ \hline
			
			D1-D2~ & 35.4 & 54.6 & 56.3 & \bf 61.7 \\
			D1-D3~ & 46.1 & 69.8 & 71.3 & \bf 77.7 \\
			D2-D3~ & 34.1 & 51.8 & 51.7 & \bf 60.1 \\
			
			\noalign{\smallskip} 
		\end{tabular}
	\end{table}
	
	We first compare in table~\ref{tab:conv4_score} the accuracy we obtained using the baseline score $S_{features}$ with different conv4 features: ResNet18 and ResNet50 trained on ImageNet, MoCo v2\cite{mocov2}, and the ResNet18 features fine-tunned by~\cite{shen2019discovering}. The feature from~\cite{shen2019discovering} achieve the best results on Physiologus manuscripts. However, on the challenging De Materia Medica manuscripts, the ResNet50 features perform best, and we thus use them in the rest of the paper.
	
	In table~\ref{tab:score}, we compare the accuracy using the different similarities explained in Section~\ref{sec:method_score}. The results obtained using $S_{trans}$ leads to the best performances on all pairs, and we consider only this score in the rest of the paper. Note in particular that optimizing the function of Equation \eqref{eqn:ransac_optimization} leads to clearly better result than using the number of inliers as in~\cite{shen2019discovering}. Since results on the Physiologus, where illustrations are fewer and more clearly different, are almost perfect, we only report the quantitative evaluation on the more challenging De Materia Merdica in the following sections.

	\begin{table*}[t]
		\caption{Accuracy of the correspondences obtained using the different normalizations explained in Section~\ref{sec:norm} and in Table~\ref{tab:eq_norm}.}
		\label{tab:normalization}
		\centering
		\scriptsize
		\begin{tabular}{c|c|c|c|c|c|c}
			\noalign{\smallskip}
			Pairs & $S$ &{$sm(\lambda S/\avg(S))$ } & {$sm(\lambda S/\max(s))$} & {$sm(\lambda S)$} & {$S/\avg(S)$} & {$S/\max(S)$} \\ \hline
			D1-D2~ & 61.7 & 68.3 & \bf 70.5 & 67.8 & 67.1 & \bf 70.5 \\
			D1-D3~ & 77.7 & 83.5 & \bf 85.3 & 83.1 & 82.5 & \bf 85.3 \\
			D2-D3~ & 60.1 & 66.3 & 66.7 & 66.1 & 65.3 & \bf 69.0 \\

		\end{tabular}
	\end{table*}
	
	\subsection{Normalization}	
	\label{sec:res_norm}
	
	In table~\ref{tab:normalization}, we compare the accuracy we obtained using the different normalizations presented in Section~\ref{sec:norm} and in Table~\ref{tab:eq_norm}. For the softmax-based normalizations that include an hyperparamter, we optimized it directly on the test data, so the associated performance should be interpreted as an upper bound. Interestingly, a simple normalization by the maximum value  $S/max(S)$ outperforms these more complexe normalization without requiring any hyper-parameter tuning. It is also interesting that all the normalization schemes we tested provide a clear boost over the raw similarity score, outlining the importance of considering the symmetry of the correspondence problem.

	\begin{figure*}[t]
		\begin{center} 
			\subfloat{\raisebox{0.08\linewidth}{\rotatebox[origin=t]{90}{Query}}}~
			\includegraphics[width=0.15\linewidth,height=0.18\linewidth]{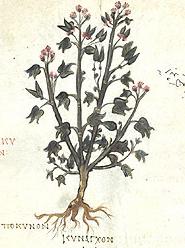}~
			\includegraphics[width=0.15\linewidth,height=0.18\linewidth]{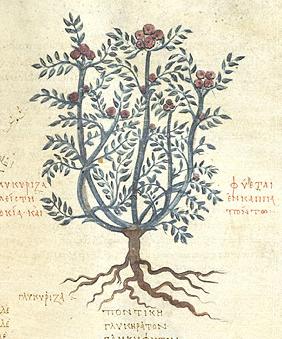}~
			\includegraphics[width=0.15\linewidth,height=0.18\linewidth]{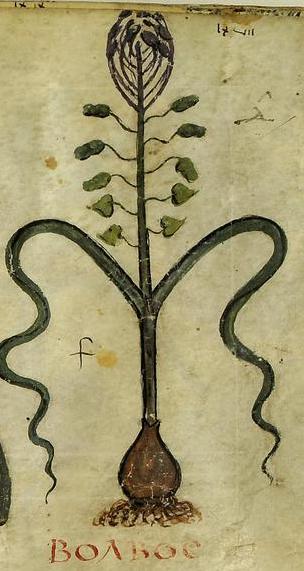}~
			\includegraphics[width=0.15\linewidth,height=0.18\linewidth]{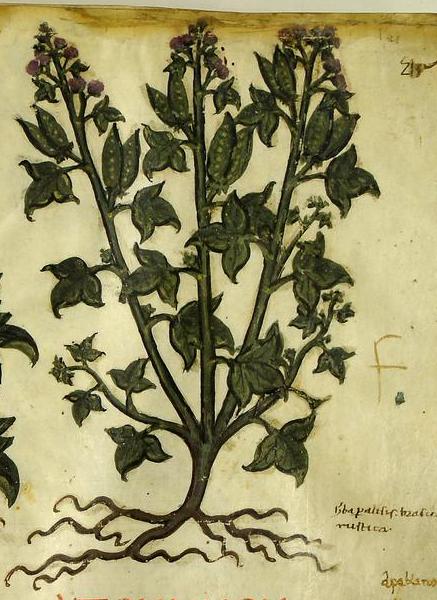}~
			\includegraphics[width=0.15\linewidth,height=0.18\linewidth]{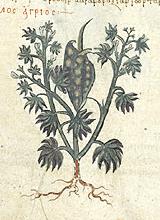}~
			\includegraphics[width=0.15\linewidth,height=0.18\linewidth]{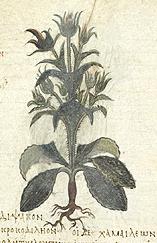}\\
			\subfloat{\raisebox{0.08\linewidth}{\rotatebox[origin=t]{90}{(Valid) Match}}}~
			\includegraphics[width=0.15\linewidth,height=0.18\linewidth]{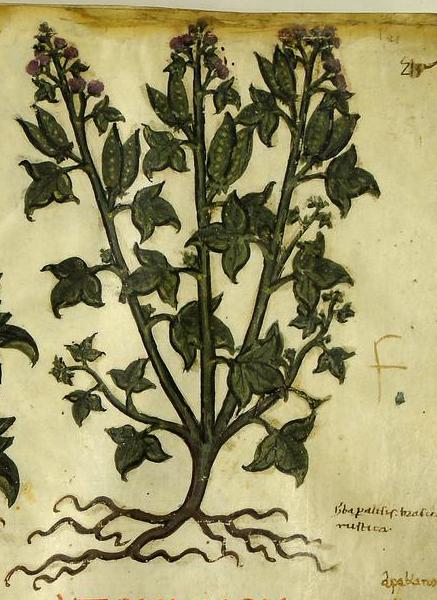}~
			\includegraphics[width=0.15\linewidth,height=0.18\linewidth]{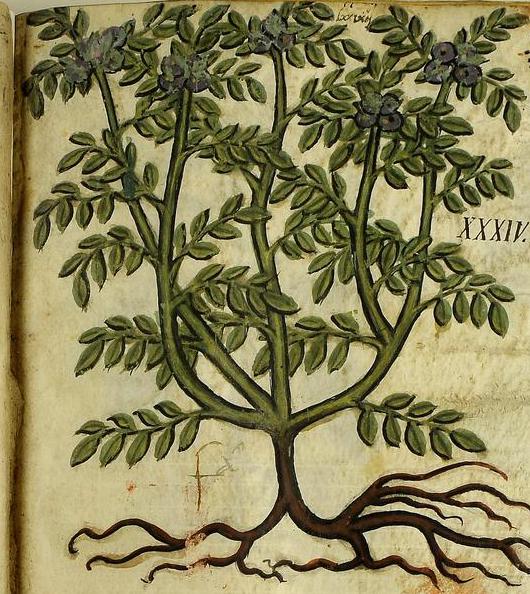}~
			\includegraphics[width=0.15\linewidth,height=0.18\linewidth]{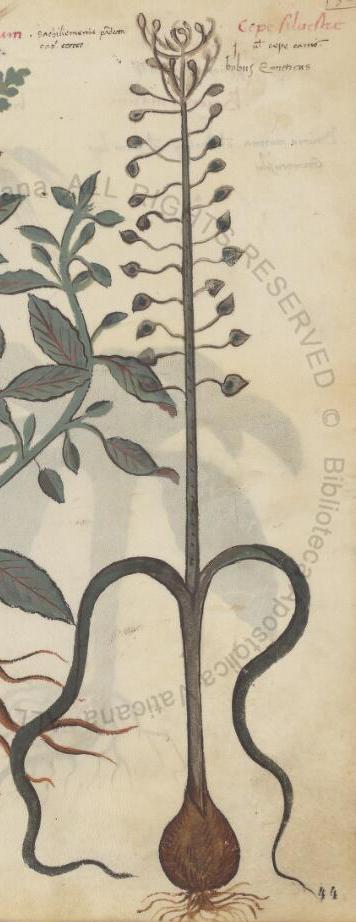}~
			\includegraphics[width=0.15\linewidth,height=0.18\linewidth]{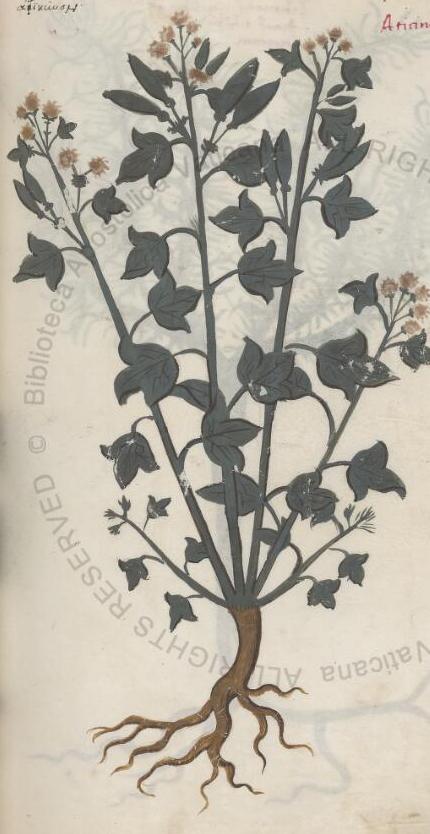}~
			\includegraphics[width=0.15\linewidth,height=0.18\linewidth]{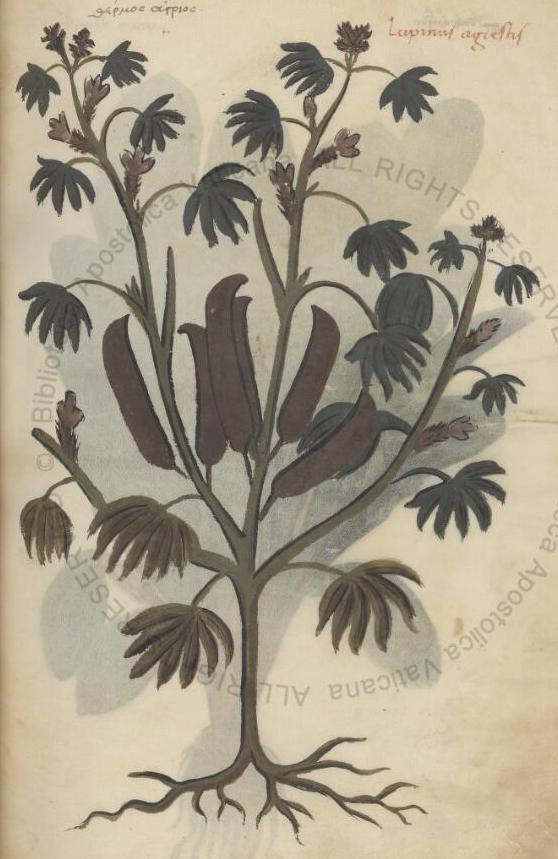}~
			\includegraphics[width=0.15\linewidth,height=0.18\linewidth]{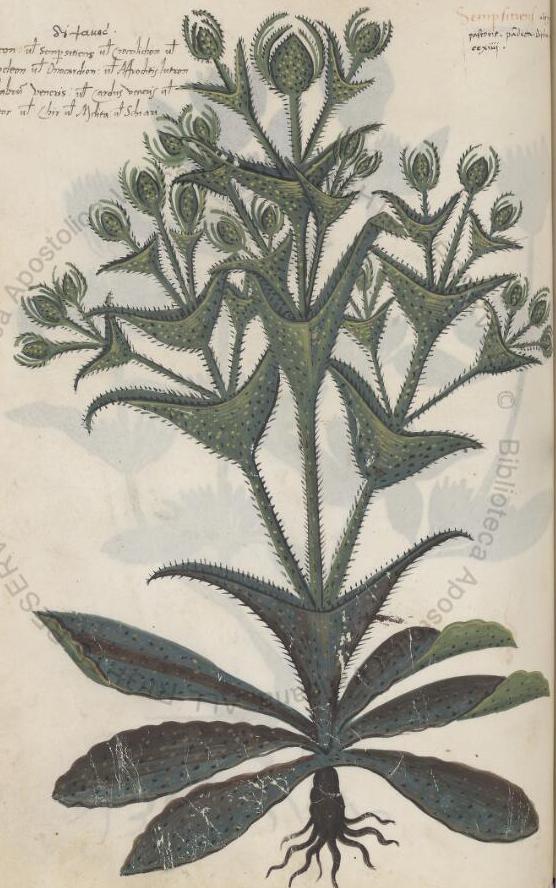}
		\end{center}   
		\caption{Examples of correspondences recovered only after the information propagation. These examples are with many local appearance changes. }
		\label{fig:diff}
	\end{figure*}

	\subsection{Information propagation}
	\label{sec:res_ip}
	
	We analyze the potential of information propagation in Table~\ref{tab:infor_prop}.
	Using the normalized similarity score, we can compute correspondences (column 'all'). Some of these correspondences will be 2-cycle or 3-cycle consistent. These correspondences will be more reliable, as can be seen in the 'only 2-cycle' and 'only 3-cycle' columns, but there will be fewer (the number of images among the annotated ones for which such a cycle consistent correspondence is found given in parenthesis). In particular, the accuracy restricted to the 3-cycle consistent correspondences is close to $100\%$. Because the accuracy of these correspondences is higher, one can use them as a set of confident correspondences $\mathcal{C}\ast$ to compute a new score $P_S$ as explained in Section~\ref{sec:prop}. The results, on all annotations, can be seen in the last two column. The results are similar when using either 2 or 3-cycle consistent correspondences for $\mathcal{C}^\ast$ and the improvement over the normalized scores is significant.

	This result is a strong evidence that important performance boost can be obtained by leveraging consistency. Qualitatively, the correspondences that are recovered are difficult cases, where the depictions have undergone significant changes, as shown in Figure~\ref{fig:diff}.
	\begin{table}[t]
		\caption{The left part of the table details the accuracy of the correspondences with the normalized score $N_S$ on different subsets of the annotated correspondences: all (the measure used in the rest of the paper), the correspondences obtained with $N_S$ are 2-cycle consistent and and the correspondences obtained with $N_S$ are 2 and 3-cycle consistent. The number in parenthesis is the number of correspondences. The right part of the table present the average accuracy obtained when performing information propagation from either the 2-cycle or the 3-cycle consistent correspondences.
		}
		\label{tab:infor_prop}
		\centering
		\scriptsize
		\begin{tabular}{c|c|c|c|c|c}
			\multirow{2}{*}{Pairs} &
			\multicolumn{3}{c|}{$N_S$} &
			{~$P_S$ -  $\mathcal{C}^\ast$:2-cycles~}& {~$P_S$ -  $\mathcal{C}^\ast$:3-cycles~}  \\
			&  all & ~only 2-cycle~ & ~only 3-cycle~ &  all & all \\
			\hline
			D1-D2~ & 70.5 {\it (295)} & 83.5 {\it (224)} & 99.2 {\it (118)} & \bf 82.5 & 82.0 \\
			D1-D3~ & 85.3 {\it (524)} & 90.2 {\it (457)} & 98.3 {\it (118)} & 88.5 & \bf 88.6 \\
			D2-D3~ & 69.0 {\it (353)} & 78.5 {\it (279)} & 96.6 {\it (118)} & \bf 81.7 & 79.3 \\
		\end{tabular}
	\end{table}
	
	\begin{figure*}[t]
		\begin{center} 
			\subfloat{\raisebox{0.08\linewidth}{\rotatebox[origin=t]{90}{Query}}}~
			\includegraphics[width=0.15\linewidth,height=0.18\linewidth]{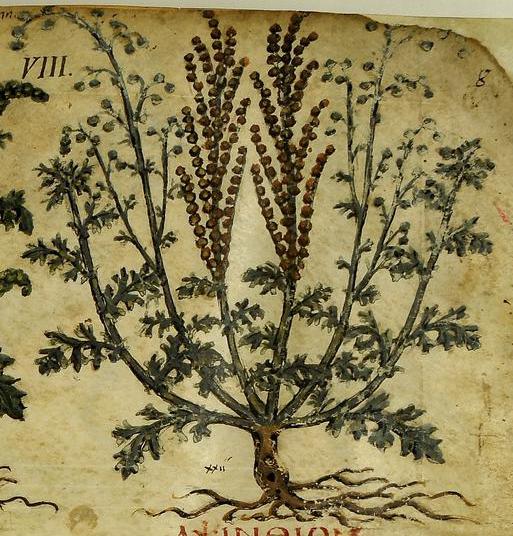}~
			\includegraphics[width=0.15\linewidth,height=0.18\linewidth]{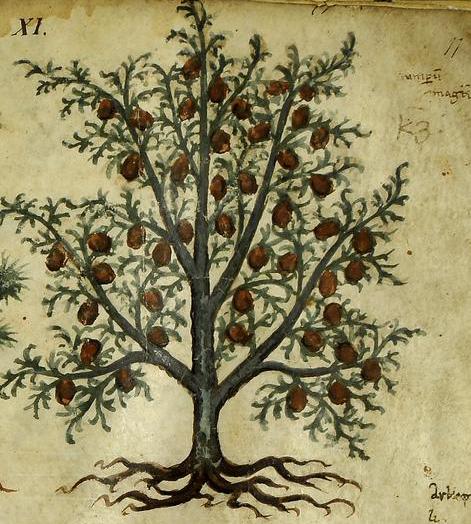}~
			\includegraphics[width=0.15\linewidth,height=0.18\linewidth]{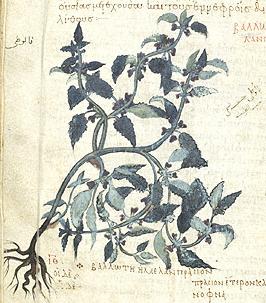}~
			\includegraphics[width=0.15\linewidth,height=0.18\linewidth]{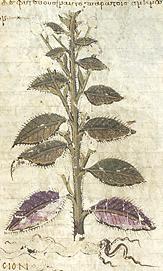}~
			\includegraphics[width=0.15\linewidth,height=0.18\linewidth]{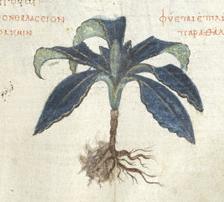}~
			\includegraphics[width=0.15\linewidth,height=0.18\linewidth]{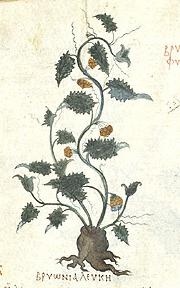}\\
			\subfloat{\raisebox{0.08\linewidth}{\rotatebox[origin=t]{90}{Predicted}}}~
			\includegraphics[width=0.15\linewidth,height=0.18\linewidth]{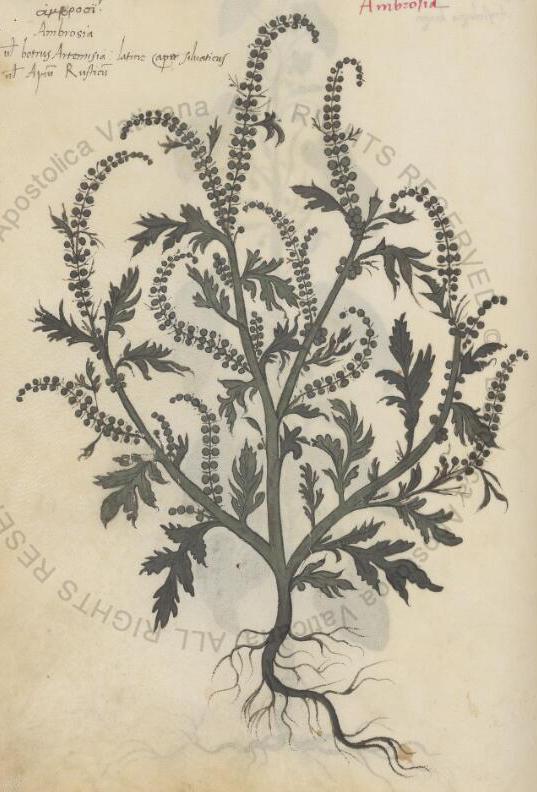}~
			\includegraphics[width=0.15\linewidth,height=0.18\linewidth]{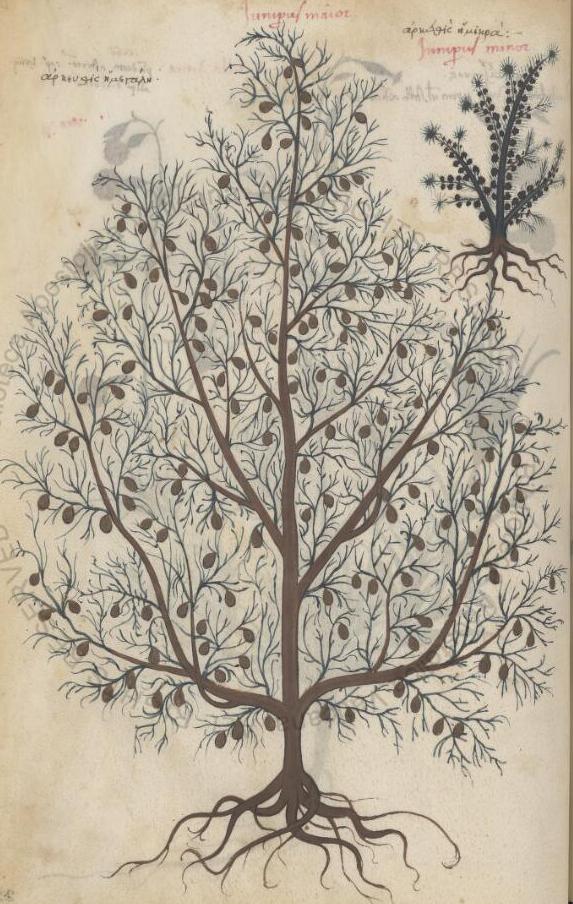}~
			\includegraphics[width=0.15\linewidth,height=0.18\linewidth]{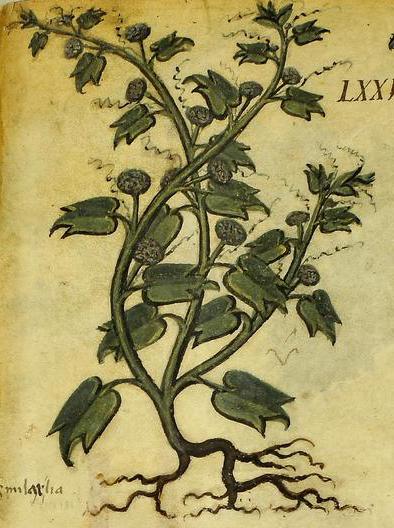}~
			\includegraphics[width=0.15\linewidth,height=0.18\linewidth]{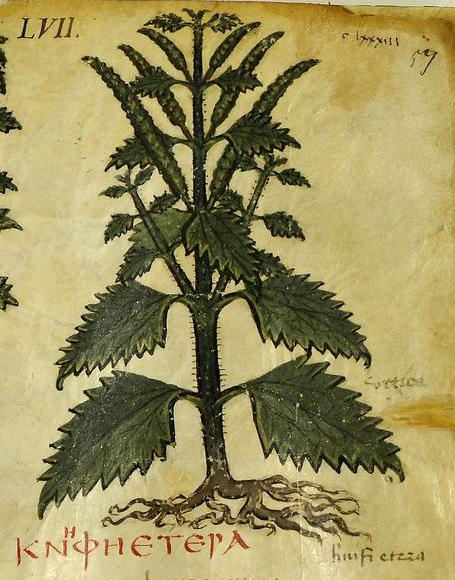}~
			\includegraphics[width=0.15\linewidth,height=0.18\linewidth]{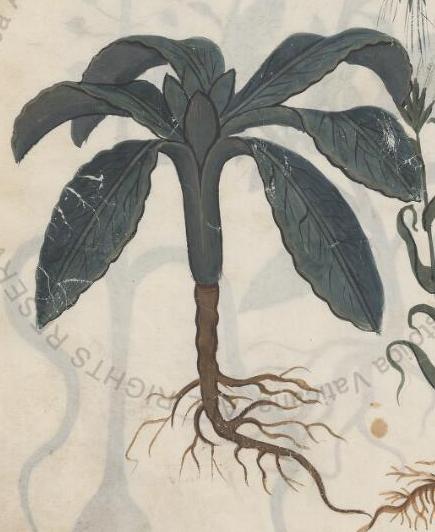}~
			\includegraphics[width=0.15\linewidth,height=0.18\linewidth]{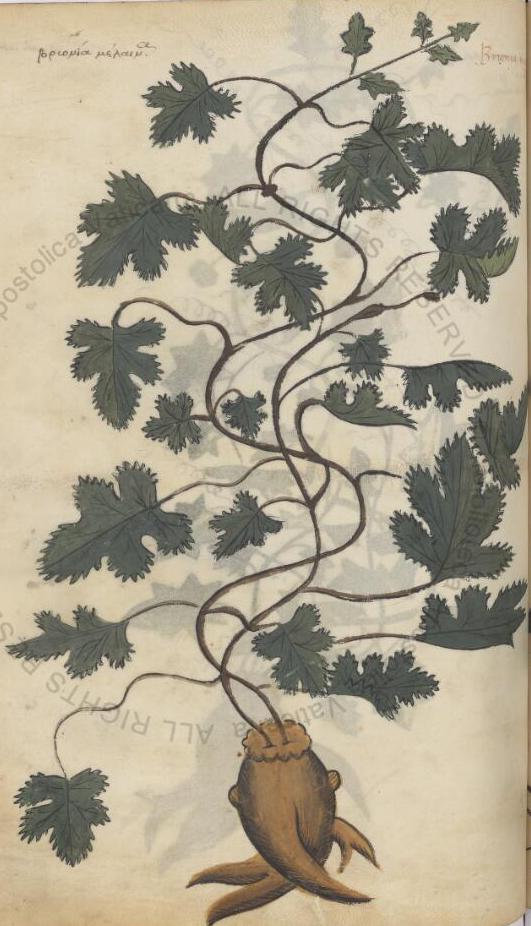}\\
			\subfloat{\raisebox{0.08\linewidth}{\rotatebox[origin=t]{90}{Ground truth}}}~
			\includegraphics[width=0.15\linewidth,height=0.18\linewidth]{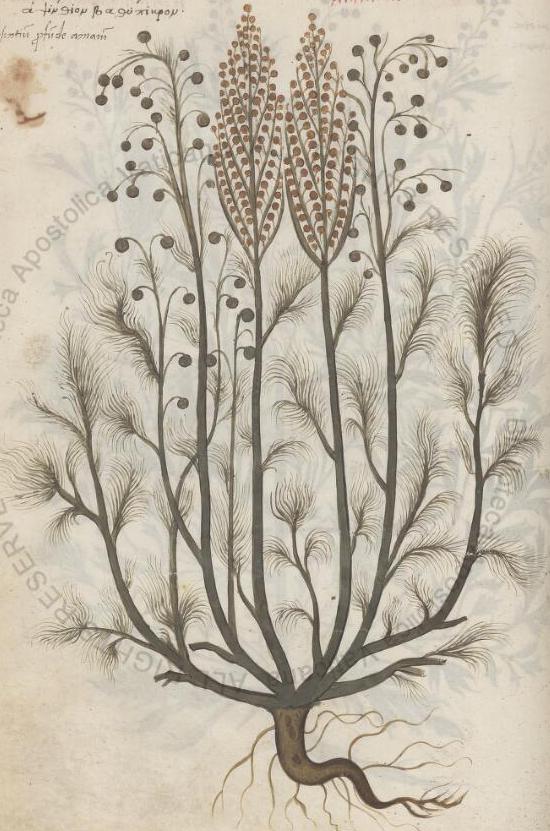}~
			\includegraphics[width=0.15\linewidth,height=0.18\linewidth]{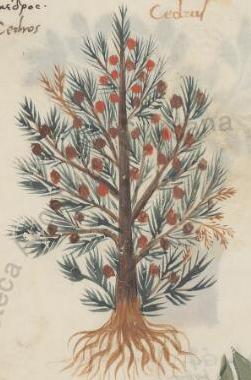}~
			\includegraphics[width=0.15\linewidth,height=0.18\linewidth]{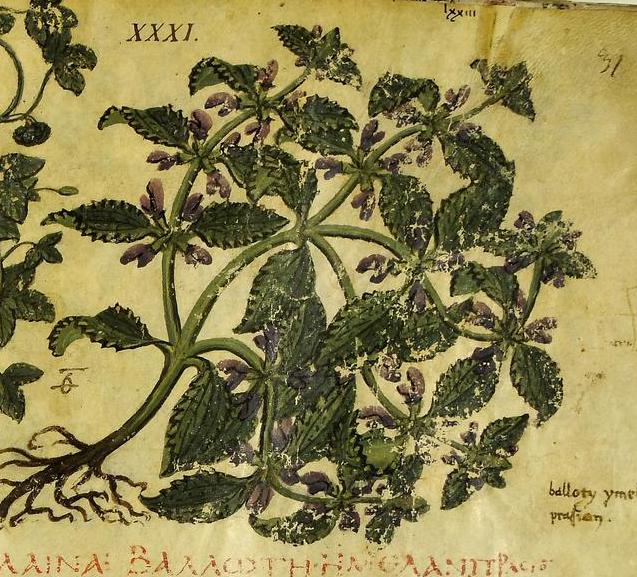}~
			\includegraphics[width=0.15\linewidth,height=0.18\linewidth]{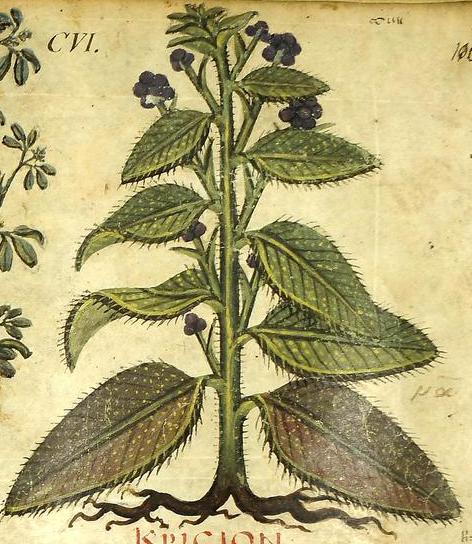}~
			\includegraphics[width=0.15\linewidth,height=0.18\linewidth]{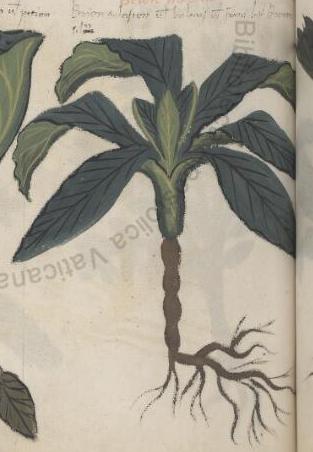}~
			\includegraphics[width=0.15\linewidth,height=0.18\linewidth]{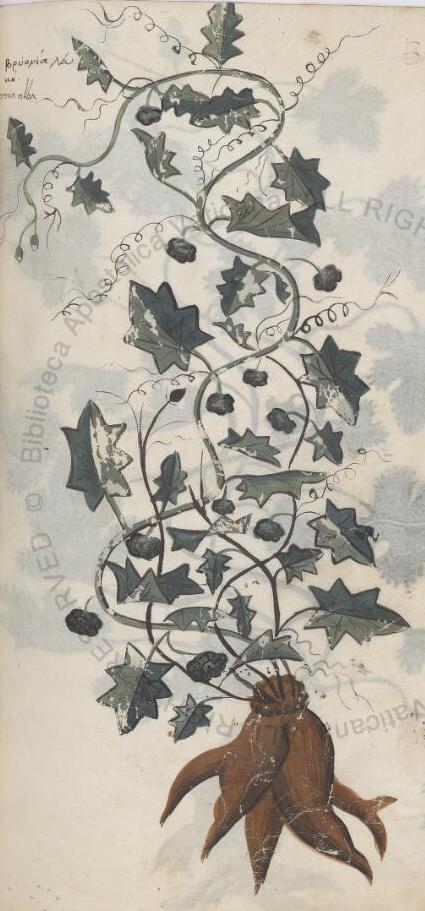}
		\end{center}   

		\caption{Examples of failure cases. We show the queries, predicted matches and ground truth correspondences in the first, second and third line respectively.}
		\label{fig:fail}
	\end{figure*}

	\subsection{Failure cases, limitations and perspectives}
	\label{sec:res_lim}
	
	Figure~\ref{fig:fail} shows some typical examples of our failure cases. As expected they correspond to cases where the content of the image has been significantly altered and where very similar images are present. For such cases, it is necessary to leverage the text to actually be able to discriminate between the images. Extracting the text and performing HTR in historical manuscripts such as ours is extremely challenging, and the text also differs considerably between the different versions. However, a joint approach considering both the text and the images could be considered and our dataset could be used for such a purpose since the full folios are available.\\

	\section*{Conclusion}
	We have introduced the new task of image collation and an associated dataset. This task is challenging and would enable to study at scale the evolution of illustrations in illuminated manuscripts. We studied how different image similarity measures perform, demonstrating that direct deep feature similarity is outperformed by a large margin by leveraging matches between local features and modeling image transformations. We also demonstrated the strong benefits of adapting the scores to the specificity of the problem and propagating information between correspondences. While our results are not perfect, they could still speed-up considerably the manual collation work, and are of practical interest.
	
	\scriptsize{
		\paragraph{Acknowledgements:}
		This work was supported by ANR project EnHerit ANR-17-CE23-0008, project Rapid Tabasco, and gifts from Adobe. We thank Alexandre Guilbaud for fruitful discussions.}
	
	\bibliographystyle{plain}
	\bibliography{sample}
	
\end{document}